\pdfoutput=1

\documentclass[11pt]{article}

\usepackage{EACL2023}

\usepackage{times}
\usepackage{latexsym}

\usepackage[T1]{fontenc}

\usepackage[utf8]{inputenc}

\usepackage{microtype}

\usepackage{booktabs}
\usepackage{graphicx}
\usepackage{todonotes}
\usepackage{caption}
\usepackage{subcaption}
\usepackage{amssymb}%
\usepackage{pifont}%
\usepackage{float}
\usepackage{xcolor}
\usepackage{listings}

\newcommand{\cmark}{\ding{51}}%
\newcommand{\xmark}{\ding{55}}%
\newcommand{\urlx}[1]{\href{#1}{#1}}

\title{The StatCan Dialogue Dataset: Retrieving Data Tables\\ through Conversations with Genuine Intents}

\author{Xing Han Lu\textsuperscript{\,1\,2$\dagger$} \hspace{3em} Siva Reddy\textsuperscript{\,1\,2\,3} \hspace{3em} Harm de Vries\textsuperscript{\,2} \\
  \textsuperscript{1\,}Mila Quebec AI Institute, McGill University \\
  \textsuperscript{2\,}ServiceNow Research \\
  \textsuperscript{3\,}Facebook CIFAR AI Chair \\
  \texttt{statcan.dialogue.dataset@mila.quebec} \\}

\date{}
\begin{document}

\renewcommand{\sectionautorefname}{Section}
\renewcommand{\subsectionautorefname}{Section}
\renewcommand{\subsubsectionautorefname}{Section}

\maketitle

\def\thefootnote{$\dagger$}\footnotetext{ Work done as visiting researcher at ServiceNow Research}\def\thefootnote{\arabic{footnote}}

\begin{abstract}
We introduce the StatCan Dialogue Dataset\footnote{ Website: \href{https://mcgill-nlp.github.io/statcan-dialogue-dataset}{mcgill-nlp.github.io/statcan-dialogue-dataset} } consisting of 19,379 conversation turns between agents working at Statistics Canada and online users looking for published data tables. The conversations stem from genuine intents, are held in English or French, and lead to agents retrieving one of over 5000 complex data tables. Based on this dataset, we propose two tasks: (1) automatic retrieval of relevant tables based on a on-going conversation, and (2) automatic generation of appropriate agent responses at each turn. We investigate the difficulty of each task by establishing strong baselines. Our experiments on a temporal data split reveal that all models struggle to generalize to future conversations, as we observe a significant drop in performance across both tasks when we move from the validation to the test set. In addition, we find that response generation models struggle to decide when to return a table. Considering that the tasks pose significant challenges to existing models, we encourage the community to develop models for our task, which can be directly used to help knowledge workers find relevant tables for live chat users.

\end{abstract}

\begin{table}[h!]
    \small
    \centering
    \begin{tabular}{l p{0.85\linewidth}}
    \toprule
    $U_1$:  & Hi, I'm looking to obtain quarterly data in regards to GDP grow (Canada), BC Housing STarts, Canada Oil Price/BBL\\
    $A_1$: & Hello, my name is Kelly C. Give me one moment as I search [...]\\ 
    $A_1$: & For GDP growth rates, please consult the following link: [...]\\
    $A_1$: & What do you mean by BC Housing Starts?\\
    \\
    $U_2$: & I'm required to research all of the housing starts for BC on a quarterly basis [...]\\
    $U_2$: & Housing starts are the number of new residential construction projects that have begun during any particular month [...]\\
    $A_2$: & I would have monthly data regarding new building permits being issued [...] \\ 
    $A_2$: & \textbf{Building permits, by type of structure and type of work: https[...]}\\
    $A_2$: & I'll have a look for oil prices. One moment.\\
    \\
    $U_3$: & Do you also have data to Canada'as oil Price/BBL ("WTI")? [...]\\
    $A_3$: & Are you looking for the retail prices of oil?\\
    $A_3$: & If so, I found some data for smaller geographies.\\
    $A_3$: & \textbf{Monthly average retail prices for gasoline and fuel oil, by geography (https[...])}\\
    $A_3$: & [...] Would those geographies be enough?\\
    $A_3$: & Or are you looking for Canada only?\\
    \\
    $U_4$: & [...] I would need something that pertains more to all of canada\\
    $A_4$: & What about this? \textbf{Monthly average retail prices for food and other selected products (https[...])}\\
    \bottomrule
    \end{tabular}
    \caption{An example of the StatCan Dialogue Dataset in which a user (U) talks to a StatCan agent (A) to find a number of data tables.  Text in \textbf{bold} indicates the title of a table retrieved by the agent.
    }
    \label{tab:selected_example_conversation}
    \vspace{-3mm}
\end{table}

\begin{table*}[t!]
    \small
    \centering
\begin{tabular}{lcccccc}
\toprule
\textbf{Dataset}                          & \textbf{Intent} & \textbf{Dialogue} & \textbf{Query}    & \textbf{Result(s)}     & \textbf{Source(s)} & \textbf{Lang.} \\
\midrule
Our work                                  & Genuine   & \cmark & Question, Request & Table link, Dial. act & StatCan & En, Fr \\
NQ \citeyearpar{kwiatkowski_natural_2019} & Mixed     & \xmark & Question          & Span excerpt           & Google, Wiki.  & En     \\
DuReader \citeyearpar{he_dureader_2018}   & Mixed     & \xmark & Question          & Span excerpt           & Baidu         & Zh     \\
OTT-QA \citeyearpar{chen_open_2020}       & Simulated & \xmark & Question          & Table/Span excerpt     & Wikipedia          & En     \\
TAPAS-NQ \citeyearpar{herzig_open_2021}   & Mixed     & \xmark & Question          & Table excerpt          & Google, Wiki.  & En     \\
CoQA \citeyearpar{reddy_coqa_2019}        & Simulated & \cmark & Question          & Span excerpt*          & Multiple      & En     \\
QuAC \citeyearpar{choi_quac_2018}         & Simulated & \cmark & Question          & Dial. act             & Wiki.          & En     \\
ATIS \citeyearpar{hemphill_atis_1990}     & Genuine   & \cmark & Request           & SQL query, Command     & TI Corp.  & En     \\
SGD-X \citeyearpar{lee2021sgd}            & Simulated & \cmark & Request           & API call, Dial. act   & Dial. Simulator   & En    \\
\bottomrule
\end{tabular}
    \caption{Comparison with related datasets (see \autoref{sec:related_works}). (*) CoQA uses rationales to support extracted answers.\vspace{-3mm}}
    \label{tab:comparison_related_datasets}
\end{table*}

\section{Introduction}
\label{sec:introduction}

One of the longstanding goals in Natural Language Processing (NLP) is to develop conversational agents that assist people with concrete tasks, such as finding information in large collections of documents or booking restaurants and hotels. To aid the development of such virtual assistants, the research community is in need of benchmarks that reflect the intents and linguistic phenomena found in real-world applications. However, developing such real-world conversational datasets is challenging in the current research landscape. On the one hand, academic labs often struggle to come up with natural use cases of task-oriented dialogue agents and collect conversations with a large number of real users. Many labs have designed artificial tasks and collected conversations from crowd workers with simulated intents~\cite{budzianowski-etal-2018-multiwoz,adlakha_topiocqa_2022,lee2021sgd}, often leading to datasets that do not capture the linguistic challenges of production settings~\cite{de_vries_towards_2020}. On the other hand, industry labs might have access to users with genuine intents (e.g., through Siri or Alexa) but rarely release such conversational datasets due to their commercial value and user privacy concerns. Hence, we argue that the research community would benefit from a task-oriented dialogue environment where findings can be validated with real users, and, to that effect, present a unique dataset in collaboration with Statistics Canada. 

Statistics Canada (StatCan) is a national statistics agency commissioned with collecting key information on Canada's economy, society, and environment. Statistics Canada conducts hundreds of surveys on virtually all aspects of Canadian life and publishes the resulting data tables on \href{https://www.statcan.gc.ca/}{statcan.gc.ca}. This website currently features 5K+ of such complex and often large data tables.  Canadian citizens---and other interested individuals---come to this website to find the statistics they are looking for. The StatCan website offers a chat functionality (available in English and French) to help users in case they can not find the appropriate information.

Sourcing from these live chats, we present the StatCan Dialogue Dataset, a collection of 20K+ English and French conversations between visitors of \href{https://www.statcan.gc.ca/}{statcan.gc.ca} and agents working at Statistics Canada. Before releasing this dataset, StatCan has ran several procedures to remove Personally Identifiable Information (PII). While we observe a wide variety of user intents, ranging from table manipulation to navigation instructions, a large number of visitors use the chat functionality to find data tables on the StatCan website. Specifically, we observe 6.6K instances where agent returns a link to a data table across 4.4K conversations. In \autoref{tab:selected_example_conversation}, we provide an example conversation in which an online user is looking for specific data tables.  

In this work, we develop two novel tasks centered on helping users find specific tables. First, we introduce the table retrieval task, which requires a model to predict the table returned by the agent given the messages sent so far. Second, we introduce the response generation task, which requires a model to predict the agent's response given the dialogue history. For both tasks, we investigate its difficulty by establishing strong baselines and evaluating them on various metrics.

We stress that both tasks are immediately useful in a real-world setting. The table retrieval task can help agents find relevant tables faster while the response generation task may lead to a virtual agent that can return relevant tables through an online conversation. We hope that this tight connection with a real-world scenario will bring the research community more insight into the challenges of developing practical dialogue agents and lead to faster transfer of research ideas and findings.

\section{Related Work}
\label{sec:related_works}

This section presents various directions related to our work. See \autoref{tab:comparison_related_datasets} for a comparative summary.

\paragraph{Open-domain QA} This is the task of answering questions using a large and diverse collection of text documents. One of the first large-scale evaluations in open-domain QA was presented at TREC-8 \citep{voorhees_trec_2001}. Since then, many studies have released large-scale open-domain QA datasets: WikiQA \citep{yang_wikiqa_2015} and MS MARCO \cite{bajaj_ms_2018} source questions from the Bing search engine, Natural Questions (NQ) \citep{kwiatkowski_natural_2019} from Google search, and DuReader \citep{he_dureader_2018} source questions in Chinese from Baidu. The questions come from real users and the answers are collected from the search results through crowd workers. Although those datasets have questions with genuine intent and the answer must be retrieved from a collection of documents, our dataset emphasizes the retrieval of tables (in a conversational setting) rather than free-form documents.

\paragraph{Table retrieval and QA}
Following works on tabular pre-training \citep{yin_tabert_2020}, table-to-text generation \citep{parikh-etal-2020-totto} and weak supervision for semantic parsing \citep{herzig_tapas_2020}, \citet{chen_open_2020} and \citet{herzig_open_2021} respectively propose OTT-QA and TAPAS-NQ, two novel approaches that extend open-domain QA to retrieving tables instead of documents. The former collects both the questions and answers from crowd workers and the latter extends Natural Questions by using tables from the article where the answer was taken. In both cases, the tables being retrieved are sourced from Wikipedia articles. Although our data also incorporate tabular retrieval, the tables are sourced from \href{https://www.statcan.gc.ca/}{statcan.gc.ca}, they can be significantly larger (as discussed in Appendix \ref{sec:appendix_formatting_size}), and they are being retrieved in an interactive and conversational setting.

\paragraph{Conversational QA}
Several works extended question answering to the conversational setting. CoQA \citep{reddy_coqa_2019} and QuAC \citep{choi_quac_2018} introduced datasets in which multiple rounds of questions are asked about a reference passage taken from a document (such as a Wikipedia article). Subsequent works extended this setup to an open-domain setting where the reference passage is not known beforehand~\cite{qu_open-retrieval_2020,anantha-etal-2021-open,adlakha_topiocqa_2022}. \citet{saeidi2018interpretation} proposed a conversational QA task about regulatory texts. Aforementioned datasets are all structured in the same way: at every turn, the first speaker will ask a question, and the other speaker will give an answer.  In contrast, the queries in our conversations are not restricted to questions, and the answers can be either a table, metadata, or a dialogue act.

\begin{table}[t]
    \centering
    \small
    \begin{tabular}{lrrrr}
    \toprule
              Dataset &  Train &  Valid &  Test &   All \\
    \midrule
              \# Conv. &   2573 &    545 &   557 &  3675 \\
              \# Turns &  11382 &   2339 &  2600 & 16321 \\
           \# Messages &  36147 &   7385 &  8340 & 51872 \\
            \# Queries &   3782 &    799 &    870 &   5451 \\
       \# Tokens / Msg &  32.83 &  33.51 & 29.32 & 32.36 \\
      \# Turns / Conv. &   4.42 &   4.29 &  4.67 &  4.44 \\
        \# Msg / Conv. &  14.05 &  13.55 & 14.97 & 14.11 \\
    \# Queries / Conv. &   1.48 &   1.47 &   1.57 &   1.49 \\
         \midrule
             \# Tables &    778 &    349 &   388 &   959 \\
         \# New tables &      0 &     41 &   145 &   181 \\
       \# Dims / Table &    3.5 &    3.5 &   3.6 &   3.6 \\
       \# Mbrs / Table &  185.5 &  210.8 & 175.6 & 172.1 \\
      \# Notes / Table &   21.1 &   22.7 &    23 &  20.4 \\
    \bottomrule
    \end{tabular}

    \caption{Statistics of English conversations and tables in the retrieval and generation tasks. New tables are calculated with respect to training set (see \autoref{tab:table_differences_and_overlaps}). \vspace{-3mm}}
    \label{tab:table_conversation_statistics}
\end{table}

\paragraph{Task-oriented Dialogue} Our work is related to work on task-oriented dialogue where users converse with virtual agents to accomplish specific goals, such as booking a restaurant or resolving a customer issue. While early work has collected a dataset in a genuine information seeking setup
~\cite{hemphill_atis_1990}, many recent datasets has collected them through a simulated setup~\cite{budzianowski-etal-2018-multiwoz, rastogi2020towards, feng2020doc2dial, Feng2021MultiDoc2DialMD, chen-etal-2021-action, lee2021sgd}. Task-oriented models usually track the dialogue state by predicting dialogue acts that are specified through intents and slot-value pairs, e.g., \texttt{findRestaurants(cuisine=Italian)}. While our dataset does not provide turn-based annotations, the released conversations come with an annotated goal i.e., which data table the user was looking for. Like other goal-oriented dialogue tasks, this annotation enables us to automatically evaluate the dialogue models through a task completion metric.   

\paragraph{Chit-chat Dialogue}
The goal for chit-chat systems is to engage in a open-ended conversation with an end-user~\cite{lowe_ubuntu_2015, dinan_wizard_2018}. Unlike our dataset, such conversations do not intend to assist the user with a specific task.

\begin{figure}[t!]
    \centering
    \includegraphics[width=0.8\linewidth]{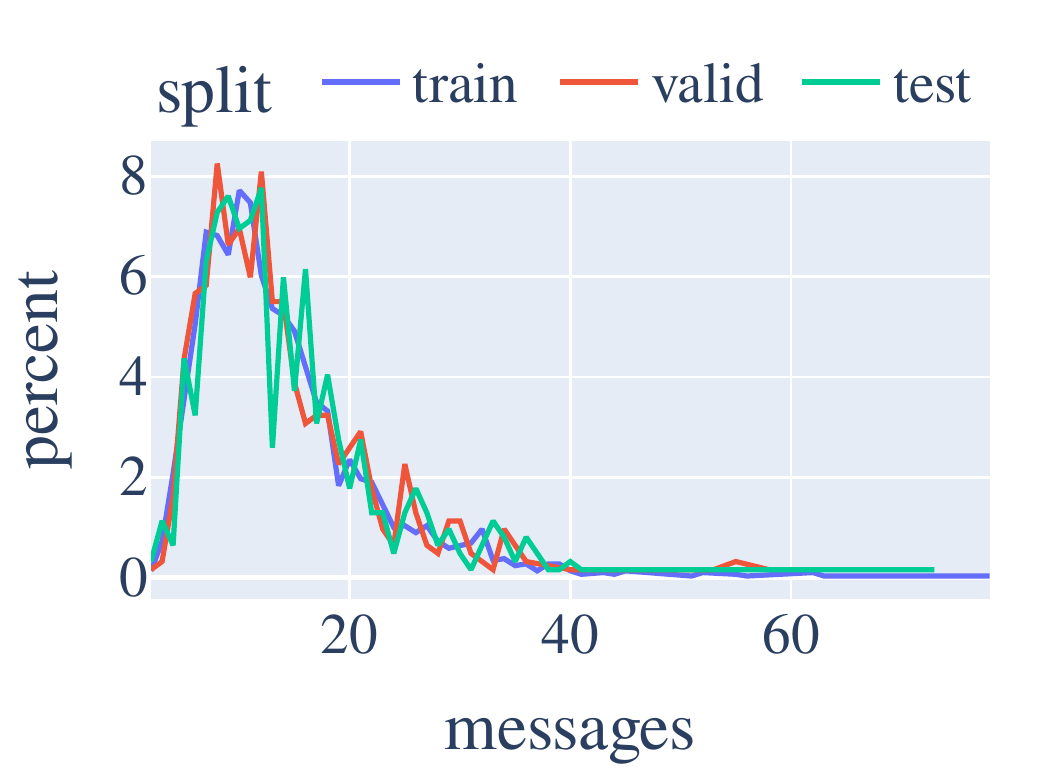}
    \caption{Histogram of messages by conversation in the both tasks (French split in \autoref{fig:turn_msg_histogram_fr}).}
    \label{fig:turn_msg_histogram}\vspace{-3mm}
\end{figure}

\section{Dataset}

The StatCan Dialogue Dataset consist of conversations collected from the live chat between March 1, 2019 till March 8, 2021. Although a variety of user intents can be found in the broader dataset of over 25K conversations, we focus on a single intent by selecting all conversations where the agent returns a data table. We use this subset to develop and test models for the two tasks that we introduce in \autoref{sec:tasks}. In \autoref{sec:basic_statistics}, we provide basic statistics about this subset of the data and present a dialogue analysis for a small number of conversations in \autoref{sec:dialogue_analysis}. In \autoref{sec:table_specs}, we turn our attention to the data tables and explain what kind of information is available for them. Finally, we explain how the dataset is split into a train, validation, and test set in \autoref{sec:conversation_selecting_splitting}. For technical specifications, a dataset card is provided in \autoref{sec:appendix_dataset_card}.

\begin{table}[t]
    \small
    \centering
    \begin{tabular}{p{0.85\linewidth} c}
        \toprule
\textbf{Merged Acts (\textit{Example})} & \textbf{\%} \\
\midrule
Answer                (\textit{You can obtain on our...}) & 50 \\
Request               (\textit{...please help me retrieve data...}) & 31 \\
Time Mgmt        (\textit{Please hold }) & 28 \\
Inform                (\textit{Please take note that...}) & 63 \\
Info Seeking Ques.  (\textit{Do you have any other...?}) & 25 \\
Promise               (\textit{...please contact the Education Ministry...}) & 18 \\
Auto Feedback          (\textit{Sure}) & 25 \\
Offer                 (\textit{...how may I help you?}) & 11 \\
Instruct              (\textit{Select at least one...}) & 18 \\
Clarif. Ques. (\textit{Which of these lines would direct...}) & 16 \\
Greeting              (\textit{Hi}) & 28 \\
Self Introduction      (\textit{My name is...}) & 17 \\
Thanking              (\textit{Thanks a lot!}) & 47 \\
Accept Thanking        (\textit{you're welcome}) & 15 \\
        \bottomrule
    \end{tabular}
    \caption{Frequency of merged speech acts occurring in 100 turns in conversations from the validation set.}
    \label{tab:merged_speech_act_counts}
\end{table}

\paragraph{User intents} The live chat was designed to fulfill specific user intents. The main intent of the chat functionality is to help users \textbf{find specific data tables}. For example, in \autoref{tab:selected_example_conversation}, the agent helps the user find tables about building permit, gasoline price, and retail prices for food. Although, users might also be interested in \textbf{obtaining meta-information}, receive help in \textbf{manipulating} a table or with the \textbf{user interface}. In some cases, the user will make \textbf{out of domain requests}. Those auxiliary intents are described in Appendix \ref{sec:userintents} since the focus of this work is on the main intent.

\paragraph{Messages and turns} Each conversation is broken down in \textbf{turns}, which is a pair of user-agent responses. Each response can have multiple \textbf{messages} sent sequentially (e.g., in \autoref{tab:selected_example_conversation}, the first agent response contains 3 back-to-back messages).

\subsection{Basic statistics}\label{sec:basic_statistics}

In total, 25397 conversations will be made available. Based on our main intent, we focus on a subset of 4468 (3675 in English and 793 in French). Out of a total of 5907 tables available in both English and French, the agents returned 959 unique tables in English 285 in French. The number of messages by conversation varies between 2 and 78 with a median of 12 for the English split (see \autoref{fig:turn_msg_histogram} for the distribution). Based on \autoref{tab:table_conversation_statistics}, there's on average 4.4 turns but 14.12 messages per conversation, with over 30 tokens for each message. This indicates that the speakers will express multiple sequential thoughts before the addressees respond. For the French split, we analyzed the basic statistics in Appendix \ref{sec:french_modeling}.

\paragraph{Frequently requested tables} In total, 6 tables make up 13.4\% of tables retrieved, covering subjects like inflation and household spending. Supplementary details can be found in Appendix \ref{sec:appendix_frequently_requested_tables}.

\subsection{Dialogue Analysis}
\label{sec:dialogue_analysis}

We categorize 100 turns (306 messages) from 24 conversations in the English validation set according to the speech acts defined by \citet{bunt_towards_2010, bunt2020iso}, which is also known as ISO standard 24617-2. We follow their taxonomy but merge some fine-grained acts with their broader concepts (e.g., \textit{correction, agreement, disagreement} with \textit{inform}). We present the speech act frequencies and examples in \autoref{tab:merged_speech_act_counts}. See the Appendix for more information on how we merged the original acts and supplementary examples (\autoref{tab:iso_speech_acts_count} and \autoref{tab:sample_linguistic_analysis}, respectively).

We notice that \textit{answers} appear twice as frequent as \textit{information seeking questions} because an interlocutor may provide an answer to both \textit{clarification questions} and \textit{requests}. Additionally, \textit{inform} acts appears 63\% of the time because agents need to expand upon their answer and users tend to clarify their initial requests by informing the other. Although less frequent, \textit{auto feedback} and \textit{time management} are still relevant because interlocutors cannot rely on visual feedback like nodding or typing. Naturally, \textit{time management} often co-occurs with \textit{promises} because the agent tends to put the user on hold while promising to fulfill their request.

\subsection{Table specifications}
\label{sec:table_specs}
\begin{table}[t]
    \small
    \centering
    \begin{tabular}{p{0.9\linewidth}}

\toprule
Title: Production and value of maple products\\
\midrule
Date range: 1924-01-01 to 2020-01-01\\
Dimensions: Geography, Maple products\\
Subject: Agriculture\\
Survey: Maple Products\\
Frequency: Annual\\
\bottomrule
    \end{tabular}
    \caption{Basic information (including title) of table in \autoref{sec:table_specs}. Full version in \autoref{tab:sample_full_info}. This can be accessed at \href{https://doi.org/10.25318/3210035401-eng}{doi.org/10.25318/3210035401-eng}.}
    \label{tab:sample_basic_info}
\end{table}

To explain the specifications, we examine a sample table with \textbf{title} \emph{Production and value of maple products} (shown in \autoref{tab:sample_basic_info}). The table has two \textbf{dimensions}, which are groups of \textbf{member items}; for \textit{geography}, the members are the provinces producing maple syrup (Quebec, Ontario, etc.), and for \textit{Maple products} the members are the production types (maple sugar, syrup, taffy and butter). A member item generalizes the concept of rows and columns as they are interchangeable via pivoting. Sometimes, details about those members are provided as \textbf{footnotes} at the end the page.

\paragraph{Basic Information} This is the core metadata and consists of the title, dimensions, subject, survey and update frequency (member items are excluded). Supplementary details are in Appendix \ref{sec:table_specs_supplementary}.

\begin{figure}[t]
    \small
    \centering
    \includegraphics[width=\linewidth]{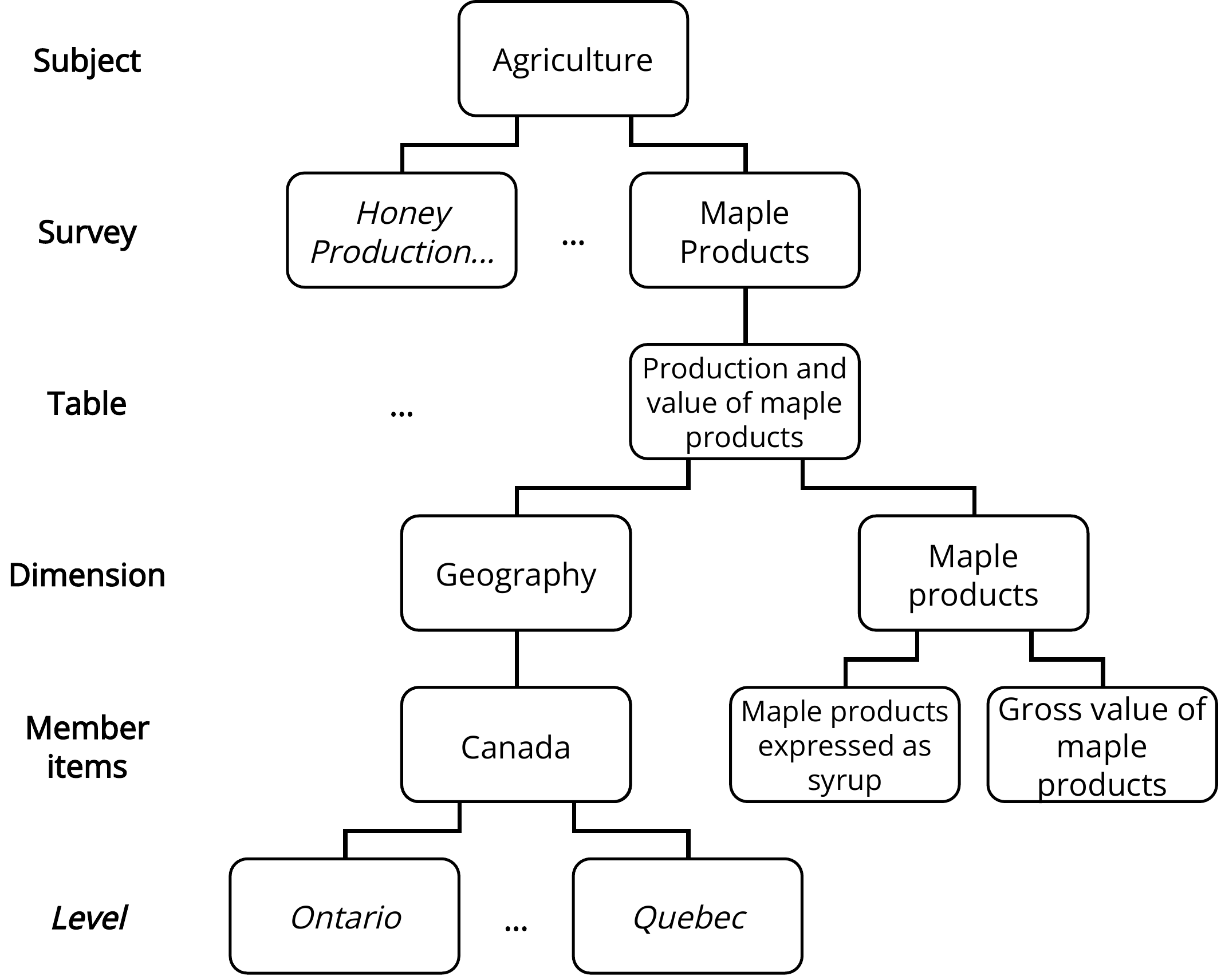}
    \caption{Diagram of the hierarchical relationship between metadata components, discussed in \autoref{sec:table_specs}.\vspace{-3mm}}
    \label{fig:diagram_hierarchical_metadata}
\end{figure}

\paragraph{Hierarchical relation} The metadata can be viewed hierarchically. As shown in \autoref{fig:diagram_hierarchical_metadata}, each subject encompasses different surveys, each survey can be used to generate one or more tables, and so on. A member item that can be nested under another member item is called \textit{Level}.

\subsection{Dataset splits}
\label{sec:conversation_selecting_splitting}
We group the conversations into a train (70\%), a validation (15\%) and a test (15\%) set. The test set was specifically selected to be the most recent conversations by date (covering Sept 8, 2020 to Mar 8, 2021), whereas the training and validation set were randomly selected from the remaining data (covering Mar 1, 2019 to Sept 8, 2020). This lets us test a model's capability to adapt to temporal shifts in the data (such as new data releases and novel events). This is useful to understand a model's capability to generalize beyond the training distribution, but it is also a better reflection of real-world applications of a model (which will be used for future data). The same splits are used for all tasks.

\section{Tasks}
\label{sec:tasks}

Based on the conversational and tabular data, we propose two tasks: (i) a table retrieval task, which requires a model to use a partial conversation to predict the table an agent will return, and (ii) a response generation task, which requires a model to use a partial conversation to generate the most probable response by the agent. The conversations in the tasks are available in both English and French.

\subsection{Retrieval task}
For this task, we truncate every conversation right before a link to a relevant table is shared by the agent. As a result, the product ID (PID) corresponding to that link becomes the objective of the retrieval task, as shown in \autoref{tab:example_retrieval_task}. When the agent shares multiple non-repeating PIDs within a conversation, each unique occurrence is treated as a separate sample.

\paragraph{Recall@k} To evaluate models for retrieval, we compute the recall at $k$ (R@k) score for $k \in \{1,10,20\}$, which corresponds to the rate where the correct table is among the $k$ tables retrieved by the model (usually ranked by a relevance score). We choose $k=1$ for real-time automatic retrieval and $k \in \{10, 20\}$ for scenarios where humans or automatic rerankers would like to use the retriever to query tables and select the best option.

\subsection{Response generation task}
\label{sec:response_generation_task}
In the first task, only the messages leading to a table retrieval are considered. For this task, each message sent by an agent is considered as a target and everything before is the source. Thus, the goal of this task is to use the source text to generate a response that matches the target (see \autoref{tab:example_generation_task}). Since dialogue responses are challenging to evaluate, we report a wide variety of metrics for this task.

\paragraph{ROUGE-L and METEOR} ROUGE-L~\cite{lin_rouge_2004, lin_automatic_2004} is a common text evaluation metric which naturally takes into account sentence level structure by identifying the longest overlapping word sequence between two sentences. METEOR~\cite{banerjee_meteor_2005} is a word-level precision and recall scoring method that encompasses different ways to represent a word, including stems and synonyms.

\paragraph{BERTScore and MoverScore} Various methods were developed to leverage contextual embeddings from BERT \citep{devlin_bert_2019} to evaluate similarity between two sentences. BERTScore \cite{zhang_bertscore_2019} computes the cosine similarity at the token level, whereas MoverScore \citep{zhao_moverscore_2019} computes the earth mover distance \citep{rubner_earth_2000} at the word or sentence level, thus capturing the cost of transforming the distribution of the generated responses into the distribution of the original responses.

\paragraph{Title accuracy} In addition to the general metrics for text generation, we also explored this task-specific metric. We define it as the proportion of generated messages that contain the title of a table shared in the reference messages. Consequently, this metric only includes turns where a table is shared by an agent. To compute this, we (i) find the product ID in the reference message, (ii) look up the title, (iii) check if that title appears exactly in the generated and reference text.

\begin{table}[t]
    \small
    \centering
    \begin{tabular}{l p{0.85\linewidth}}
    \toprule
       & \textbf{Source text (on-going conversation)} \\
       & [...]\\
    A: & What do you mean by BC Housing Starts? \\
    \\
    U: & I'm required to research all of the housing starts for BC on a quarterly basis[...]\\
    U: & Housing starts are the number of new residential construction projects that have begun during any particular month\\
    A: & [...] I would have monthly data regarding new building permits being issued.\\
    \midrule
       & \textbf{Retrieval target (StatCan table)}\\
       & \textit{Table 34-10-0066} (Building permits, by type of structure and type of work)\\
    \bottomrule
    \end{tabular}
    \caption{Source and targets of the retrieval task, based on \autoref{tab:selected_example_conversation}. Given the on-going conversation, the goal is to retrieve a StatCan table. \vspace{-2mm}
    }
    \label{tab:example_retrieval_task}
\end{table}

\section{Models}
\label{sec:models}

To help understand the performance of finetuned models on our tasks, this section presents an overview of the methods for the English splits, whereas the implementation details are covered in Appendix \ref{sec:appendix_implementation_details}. Similar architectures were used for French (described in Appendix \ref{sec:french_modeling}).

\subsection{Retrieval}

\paragraph{BM25} We use \citet{robertson_probabilistic_2009}'s algorithm to retrieve the metadata of a table (passage) similar to a given query by weighting the \textit{idf}-scaled term frequency of query words with respect to the passages. 

\paragraph{DPR} Proposed by \citet{karpukhin_dense_2020}, Dense Passage Retrieval (DPR) is a pair of transformer models that separately encode a query and a passage, and the dot product of the resulting vectors will have a higher score if the passage is relevant to the query. We finetune this model to retrieve the metadata of a table (passage) given the on-going conversation (query).

\begin{table}[t]
    \small
    \centering
    \begin{tabular}{l p{0.85\linewidth}}
    \toprule
        & \textbf{Source text (on-going conversation)} \\
    U:  & Hi, I'm looking to obtain quarterly data in regards to GDP grow (Canada), BC Housing STarts, Canada Oil Price/BBL\\
    A: & Hello, my name is Kelly C. Give me one moment as I search [...]\\
    A: & For GDP growth rates, please consult [...]\\
    U: & I'm required to research all of the housing starts for BC on a quarterly basis [...]\\
    \midrule
       & \textbf{Generation target (next response by agent)} \\
    A: & I would have monthly data regarding new building permits being issued. [...]\\
    \bottomrule
    \end{tabular}
    \caption{Source and targets of the response generation task, based on \autoref{tab:selected_example_conversation}. Given the on-going conversation, the goal is to generate the agent's response. \vspace{-2mm}
    }
    \label{tab:example_generation_task}
\end{table}

\paragraph{TAPAS and TAPAS-NQ} \citet{herzig_tapas_2020} introduced a model that learned to encode flattened tables cells in a self-supervised manner during pre-training. We finetuned it to retrieve the truncated content of a table given an on-going conversation. Subsequently, \citet{herzig_open_2021} finetuned TAPAS to perform open-domain table retrieval on 12K questions-answer-table triplets extracted from NQ; we further finetune this variant in the same way and report the results as TAPAS-NQ.

\paragraph{Exploring table representation} In the simplest scenario, only the title is given to BM25 and DPR. Moreover, we evaluate variants that encode the basic information, member items, footnotes, or a combination of them. For TAPAS and TAPAS-NQ, we also finetuned a variant that retrieves the title, dimensions and member items, since the original TAPAS could attend titles and column names.

\subsection{Response generation}

\paragraph{T5} We finetuned the large variant of T5 \citep{raffel_exploring_2020} (named \textit{No aug.} in \autoref{tab:selected_generation_results_english}) to auto-regressively decode the target (agent reply) after first encoding the source (on-going conversation).

\paragraph{Augmenting T5 with top-$k$ title(s)} For every partial conversation, we use DPR (basic+member) to retrieve the top-$k$ tables (where $k \in \{1, 5\}$), and append their titles to the partial conversation. This allows T5 to decide between using one of the suggested titles and generating something else (e.g., clarification question). This is similar to the agents' behavior, as they tend to return a title with the URL when sharing a relevant table. Furthermore, supervising T5 to ignore or return a title is equivalent to an implicit binary classification.

\begin{table}[t]
    \small
    \centering
\begin{tabular}{llll}
\toprule
    Metadata &     R@1 &    R@10 &    R@20 \\
\midrule
Basic & 14.7 &  45.0 &  55.0 \\
Basic + member & 15.7 &  46.2 &  56.3 \\
Basic + footnotes & 13.9 &  44.4 &  54.2 \\
Member &  10.7 &  35.0 &  46.3 \\
Title & 13.9 &  43.8 &  53.4 \\
\bottomrule
\end{tabular}
    \caption{Retrieval results of DPR for the English test split with varying table representations. Overview of metadata in \autoref{sec:table_specs}. %
    \vspace{-2mm}}
    \label{tab:selected_retrieval_results_english}
\end{table}
\begin{table}[t]
    \small
    \centering
\begin{tabular}{llll}
\toprule
    Model &     R@1 &    R@10 &    R@20 \\
\midrule
BM25 & 0.3 & 2.3 & 3.8 \\
DPR & 14.3 & 45.1 & 54.2 \\
TAPAS &  6.1 &  22.1 & 31.5 \\
TAPAS-NQ & 7.4 & 30.0 & 39.3 \\
\bottomrule
\end{tabular}
    \caption{Retrieval results for the English test split when encoding title and member items. DPR and TAPAS were run 3 times and averaged. 
    \vspace{-2mm}}
    \label{tab:retrieval_results_varying_representations}
\end{table}

\begin{table}[t]
    \small
    \centering
\begin{tabular}{lrrr}
\toprule
Metrics &   No aug. &  Top-1 Title &  Top-5 Titles \\
\midrule
METEOR     &  23.35 &        24.07 &         24.41 \\
ROUGE-L    &  30.65 &        30.76 &         30.88 \\
MoverScore &  59.82 &        60.23 &         60.31 \\
BERTScore  &  86.04 &        86.11 &         86.17 \\
Title Acc. &   6.96 &         7.99 &         10.82 \\
\bottomrule
\end{tabular}
    \caption{Response generation results for the English test split.\vspace{-3mm}}
    \label{tab:selected_generation_results_english}
\end{table}

\section{Results and Discussions}
\label{sec:results_analysis}

Based on our baselines and data, we report the results and analyze the challenges that our dataset and tasks pose for existing models. For the English splits, the main retrieval results are reported in \autoref{tab:retrieval_results_varying_representations} and \autoref{tab:selected_retrieval_results_english}, and main generation results are in \autoref{tab:selected_generation_results_english}. Full results can be found in Appendix \ref{sec:appendix}, respectively in \autoref{tab:full_retrieval_results_english} and \autoref{tab:full_generation_results}, and relevant statistical tests in Appendix \ref{sec:appendix_stats_tests}.

\paragraph{Impact of table representation}
In \autoref{tab:retrieval_results_varying_representations}, we observe that the metadata representation affects the retrieval recall. Although DPR can achieves respectable results when it only retrieves the title, including basic information (defined in  \autoref{sec:table_specs}) yields slight improvements, and further adding member items results in a significant difference from only using title ($p=0.014$). However, only using member item result in drastic decrease in recall ($p=0.00086$), indicating the importance of the title. Moreover, footnotes do not yield any improvement, which may be because they often exceed the maximum context span (see \autoref{tab:sample_full_info}). Thus, concisely but meaningfully representing metadata will be crucial to achieve a good recall on the retrieval task.

\paragraph{Transfer to table retrieval task}
Our experiments allow us to analyze the effectiveness of open-domain QA  fine-tuning (NQ) and tabular pre-training when transferring to our table retrieval task. We observe in \autoref{tab:retrieval_results_varying_representations} that DPR outperforms TAPAS and TAPAS-NQ by respectively 23.0\% and 15.1\% in test recall@10. Moreover, TAPAS-NQ achieves a better performance when it only retrieves the title and member items instead of the full table ($p=0.016$), likely due to repetitions and truncation due to context size limits. Although both DPR and TAPAS-NQ were trained on NQ, the latter was trained on a small subset (12K vs 320K) that contains tables. Our experiments indicate that TAPAS transfers poorly from one task (NQ-Tables) to another (StatCan).

\paragraph{Response generation}
We compare the performance of fine-tuned T5-large models with and without DPR-augmented table titles.  In \autoref{tab:selected_generation_results_english}, we notice that retrieval-augmented models show modest improvements on the 4 non-task specific metrics. However, the top-5 augmented model achieves an absolute improvement of 3.86\% in title accuracy, indicating that the information provided by DPR does help T5 in generating the desired title. It is nevertheless surprising that T5 without augmentation achieves a score of 6.96\%, suggesting that the T5 model is capable of storing the titles seen at training time, and, to a limited extent, is able to recall and return them at test time. Finally, we point out that the title accuracy is still 5.91\% lower than top-1 recall of the DPR retriever (\autoref{tab:selected_retrieval_results_english}), indicating that T5 fails to learn when to return a table (despite the agent retrieving a table in 23.4\% of all turns). In the case of top-5 titles, T5 struggles to decide which table title to return.

\paragraph{Qualitative analysis of generated responses}
We examine various conversations to understand what type of responses are generated by T5. We find that it can generate simple speech acts like greetings, but can struggle with context-specific speech acts such as clarification questions. Moreover, it can reliably reply with the title of a common table, struggles for uncommon ones, and is sometimes capable of generating unseen titles with the help of DPR. The full analysis is in Appendix \ref{sec:appendix_case_study_t5}.

\begin{figure}[t]
    \small
    \centering
    \includegraphics[width=0.75\linewidth]{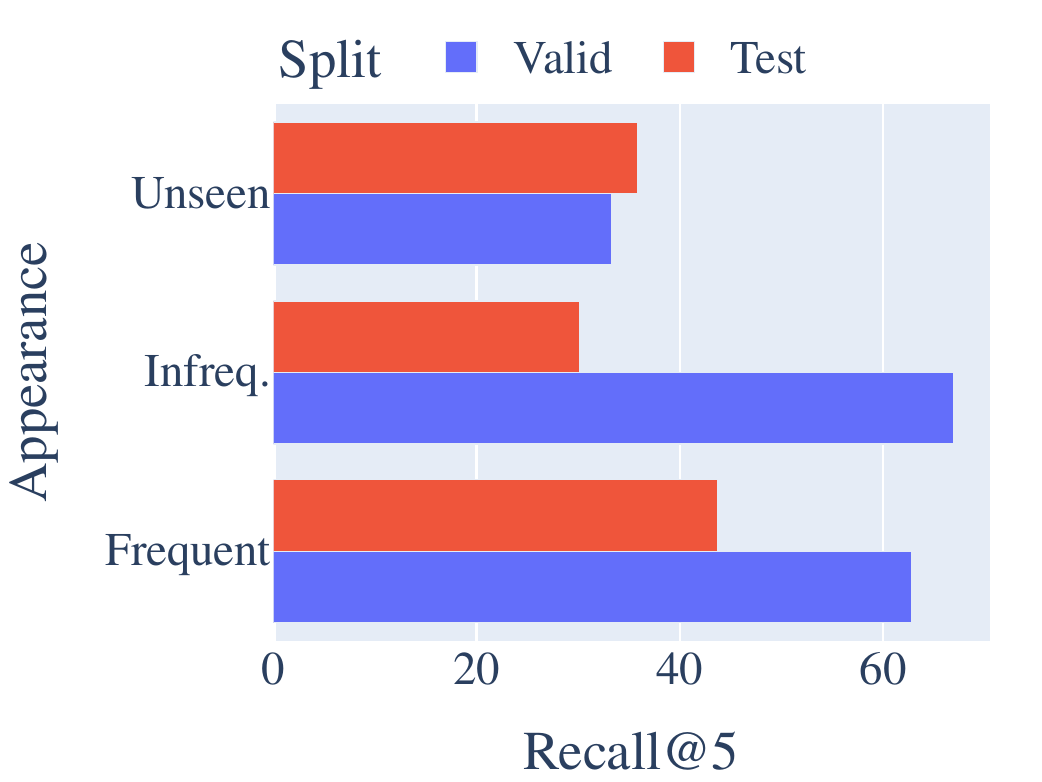}
    \caption{Results for DPR B+M for tables appearing in the training set frequently (10+ times), infrequently (1-10 times), and unseen at train time. \vspace{-3mm}}
    \label{fig:recall_at_5_by_appearances}
\end{figure}

\begin{figure}[t]
    \small
    \centering
    \includegraphics[width=0.75\linewidth]{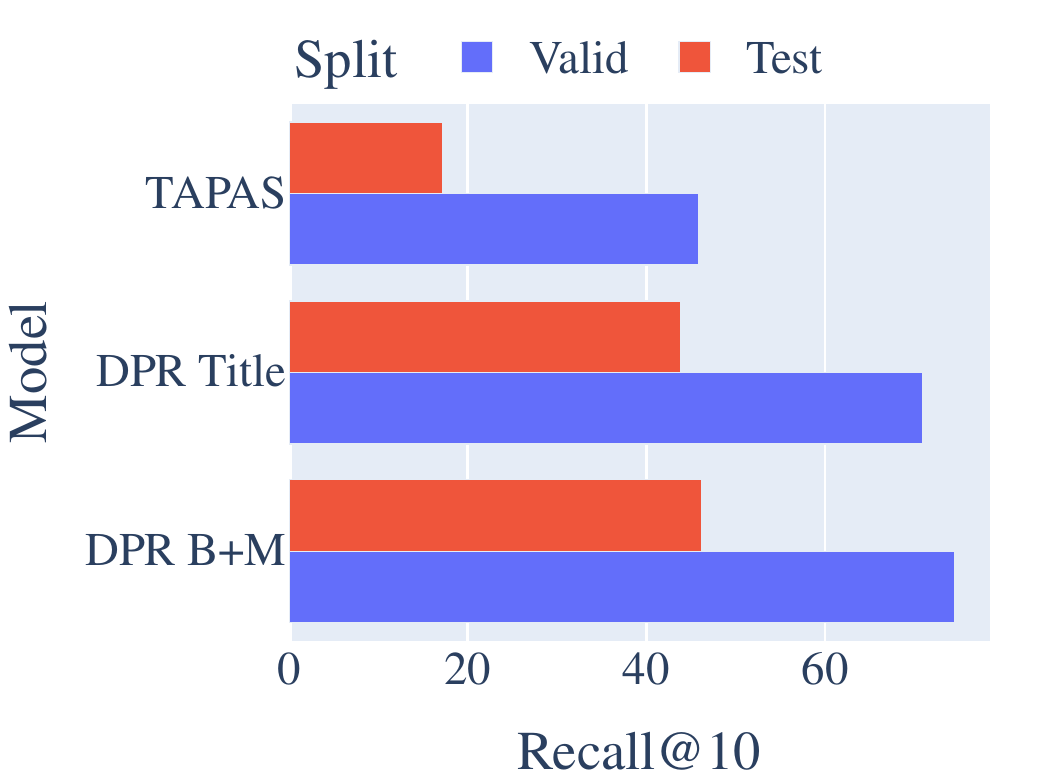}
    \caption{Validation and test recall for a selected set of retrieval models. We observe a significant drop in performance. \textit{B+M} denotes Basic + member. \vspace{-3mm}}
    \label{fig:valid_vs_test_acc}
\end{figure}

\paragraph{Temporal drifts}
\label{par:temporal_drift}
As explained in \autoref{sec:conversation_selecting_splitting}, we use a temporal split to test the model's ability to generalize to future conversations. We observe a significant drop in recall (13\%-28.3\%) in \autoref{fig:valid_vs_test_acc} when we compare the validation and test set performance, even when the models are trained with varying metadata representations. Similarly, T5 achieves low scores on the test split for the response generation task (\autoref{tab:selected_generation_results_english}). This large gap suggests that trained models struggle to generalize to future conversations. First, we found that this is likely caused by the number of new tables that appear in the test split (145) compared to validation (41), as shown in \autoref{tab:table_conversation_statistics}. Moreover, the subjects of the conversations have significantly changed: users started to care more about businesses, health and IT, and less about demography, income and pensions. This is likely motivated by real-world events affecting the users, which are more difficult to implicitly capture from simulated environments, but desirable in order to understand a model's robustness in temporal shift and for real world applications. In the Appendix, \autoref{fig:category_distribution} displays the differences between the training and test splits for all subjects.

\paragraph{Generalizing to unseen tables}
\label{par:generalizing_unseen_tables}
As shown in \autoref{fig:recall_at_5_by_appearances}, DPR performs well for tables appearing frequently in the validation split, but poorly in the test split, which could be caused by temporal drift. As expected, tables that were not seen during training resulted in poor recall@5 in either splits. Moreover, the difference in recall between valid and test for infrequent tables could be caused by many potential reasons (learning bias, temporal overfitting, spurious correlation with hidden factors). Thus, future models should aim to close the gap between unseen and frequent tables and within the temporal spectrum of infrequent tables.

\paragraph{French results}
In both tasks, we see a drop across all metrics for all models, likely due to the smaller dataset size.  Some observations remain valid: temporal drift, poor BM25 performance, and augmentations benefit mT5 for certain metrics. However, others differ: adding member items hurts test results and mT5 performs poorly on \textit{title accuracy}. Modeling details and results can be found in the Appendix \ref{sec:french_modeling}. 

\section{Conclusion}
In this paper, we introduce the StatCan Dialogue Dataset, a novel corpus consisting of 20K+ English and French conversations between online visitors of \href{https://www.statcan.gc.ca/}{statcan.gc.ca} and operators of Statistics Canada. Based on this dataset, we propose two tasks centered on helping users find specific data tables: the table retrieval task and the response generation task. For the table retrieval task, we experiment with various DPR and TAPAS variants, finding that DPR strongly outperforms its TAPAS counterpart, as well as the BM25 baseline. For the response generation task, we investigate fine-tuned T5-large models and explore variants where the input is augmented with table titles from DPR. We find that retrieval-augmented T5 models more frequently return the correct tables, although its title accuracy is still lower than the corresponding recall of the DPR retriever. This result suggests that the generation models struggle to decide when to return a table. We also find that retrieval and generation have difficulty generalizing to future conversations, as our temporal test split revealed a big performance gap between the validation and test set. All in all, we believe that our tasks pose significant challenges to currently available models and encourage the research community to further explore this dataset and build conversational models that help users of Statistics Canada.

\section*{Limitations}
\label{sec:limitations}

\paragraph{Tasks and models limitations} 
The tables in the retrieval task are sourced from \urlx{statcan.gc.ca}, which means that the content is primarily about Canadian demographics\footnote{More information can be found here: \url{https://www.statcan.gc.ca/en/subjects-start/population_and_demography}} and are professionally edited by StatCan employees. Moreover, the generation task is specifically designed to model responses with high fidelity based on retrieved tables, so this task should not be directly used in an unintended or non-research setting (e.g., deploying a virtual assistant) as they pose risks of hallucination that could negatively impact stakeholders. Furthermore, those limitations can be reflected in the models we trained, so we will share those limitations in the model cards \citep{mitchell_model_2019} on release.

\paragraph{Environment impact}
We acknowledge the models in \autoref{sec:models} used hardware with significant energy consumption. We purposefully chose models of reasonable sizes that can be reproduced on one GPU. Additionally, our hardware is powered by renewable energy. %

\paragraph{Artifacts and computational experiments}
We trained models using libraries based on their intended use and we will release the relevant artifacts following the original licenses. The computational details of the experiments are described in Appendix \ref{sec:appendix_implementation_details}.

\section*{Ethics Statement}

\paragraph{Privacy and data access} 
As discussed in \autoref{sec:introduction}, significant efforts were made to remove Personally Identifiable Information (PII). However, we do not rule out the possibility that certain details could have been missed in that process. Thus, any user that wishes to use the data will need to authenticate and accept the terms of use through an institutional data repository; the terms will require the user to report any instance of PII leak, which will be removed with a dataset update. Additionally, we request any derivative or modifications to be published in the same data repository with the original terms of use and licenses preserved or extended.

\paragraph{Risk of toxicity in online discourse}
StatCan agents are trained to work with online users in a professional manner. Moreover, since the users access \urlx{statcan.gc.ca} anonymously and virtually, it is more likely to observe toxic online disinhibition \citep{lapidot-lefler_effects_2012}, which could translate to toxicity in users' utterances. Thus, we request dataset users to report any instance of toxicity in conversations, which will be reviewed in the same manner as PII leaks. 

\section*{Acknowledgement}
We thank our collaborators at Statistics Canada for providing us the dataset, guiding us through the technical aspects of the tables, and sharing valuable feedback on the project. We thank Sivan Milton for the helpful discussions on dialogue acts and analysis.

\bibliographystyle{acl_natbib}
\bibliography{bibliography}

\appendix

\section{Appendices}
\label{sec:appendix}

\subsection{Complete User Intents}
\label{sec:userintents}

To provide insight into what kind of help is offered by StatCan's live chat, we qualitatively analyze the conversations and highlight examples of the main user intents below. 

\paragraph{Finding a table} The main intent of the chat functionality is to help users find specific data tables. For example, one user was looking for the  population numbers in certain regions of Montreal for 2012-2016. This intent is the focus of our work.

\paragraph{Obtaining meta-information of table} Instead of finding data tables some users are interested in meta information of a specific table. For example, one visitor wanted to know when the next Census is released. Another user was interested in understanding the definition of Workforce Availability (WFA) and Labour Market Availability (LMA).

\paragraph{Manipulating a data table} Some users would like to obtain the data tables in a different format or representation. For example, one user was looking at a specific data table and asked if they can see annual instead of monthly values. 

\paragraph{Help with user interface} Some users are looking for help with the user interface. For example, one user wanted to download a specific data table but they were unable to find the download link. 

\paragraph{Out of domain requests} We find many conversations that are outside of the scope of StatCan's live chat. For example, some user asked what documentation needs to be provided to ship a specific product to a foreign country. 
\newline

The first intent is covered in \autoref{tab:selected_example_conversation}, and subsequent intents are in \autoref{tab:examples_user_intents}. While we believe all intents are interesting directions for dialogue research, we focus on the table retrieval intent because (i) there are many conversations available for them and (ii) there is a clear measure of task success i.e., whether the correct table is retrieved. Throughout the rest of this paper, we work with conversations where the agent returns a table URL.

\subsection{Supplementary Table Description}

\subsubsection{Formatting and size}
\label{sec:appendix_formatting_size}
Full tables are stored in a \textit{long} format, where each column corresponds to a dimension, except the last one which corresponds to the value. The number of rows corresponds to the Cartesian product of the dimensions' members. This means that for $n_i$ members across $d$ dimensions, we have a total of $\prod_{i=0}^{d} n_i$ rows in the full table. When a table has many member items, the full table can become extremely large. Moreover, tables shown on the statcan.gc.ca are usually a pivoted and filtered view of the full table, which means certain members will become columns, others will become rows, and many are simply omitted.

\subsubsection{Detailed Specifications}
\label{sec:table_specs_supplementary}

This section provides supplementary details for \autoref{sec:table_specs}.

\paragraph{Product ID (PID)} Unique 8 to 10-digit identifier given to each published data table. Although other types of tables might be shared by the agent, the PID will always be given for the official data tables; as a result, any table that does not have a PID in the URL is not considered for this task. The first two digits (1-2) represent the code of the subject associated with the table (this can be found in the basic information), then 3-4 represent the product type, which in our case are tables and are common coded as ``10". Digits 4-8 is a unique identifier representing that table for the given subject. Digits 9-10 indicate the view of the table and will commonly be ``01", which represents the default view; they are needed when constructing the URL but they are otherwise optional, and are omitted in the released dataset.

\paragraph{Member items} Labels for individual tables, and can either be represented as a column or a row index through a pivot operation. Each table will have member items different from other tables.

\paragraph{Dimensions} Non-overlapping sets of member items; each member item must belong to a dimension. For a table with $d$ dimensions, each data value is associated with a single $d$-tuple in the $d$-fold Cartesian product of the dimensions, but not every tuple in the product will have an associated data value (if it was not recorded or if the tuple is invalid).

\paragraph{Basic information} Metadatum consisting of the title, the date range, the frequency, the dimensions, the subject category, and the survey\footnote{ The list of surveys and link to detailed information can be found at \href{https://www.statcan.gc.ca/en/survey/list}{https://www.statcan.gc.ca/en/survey/list} } from which the data was sourced.

\paragraph{Footnotes} Unstructured comments often included with a table if supplementary details need to be given; usually, those notes are associated with a specific member item that requires more explanations.

\paragraph{Full table} For all tables with a PID, the full tables (as a CSV file), their complete metadata, and their basic information are made available as part of the data release and can be used in our proposed tasks. A sample table can be found in \autoref{sec:table_specs}.

\subsubsection{Table updates and archives}
\label{sec:appendix_table_updates}

\begin{figure*}[h]
    \small
    \centering
    \includegraphics[width=\linewidth]{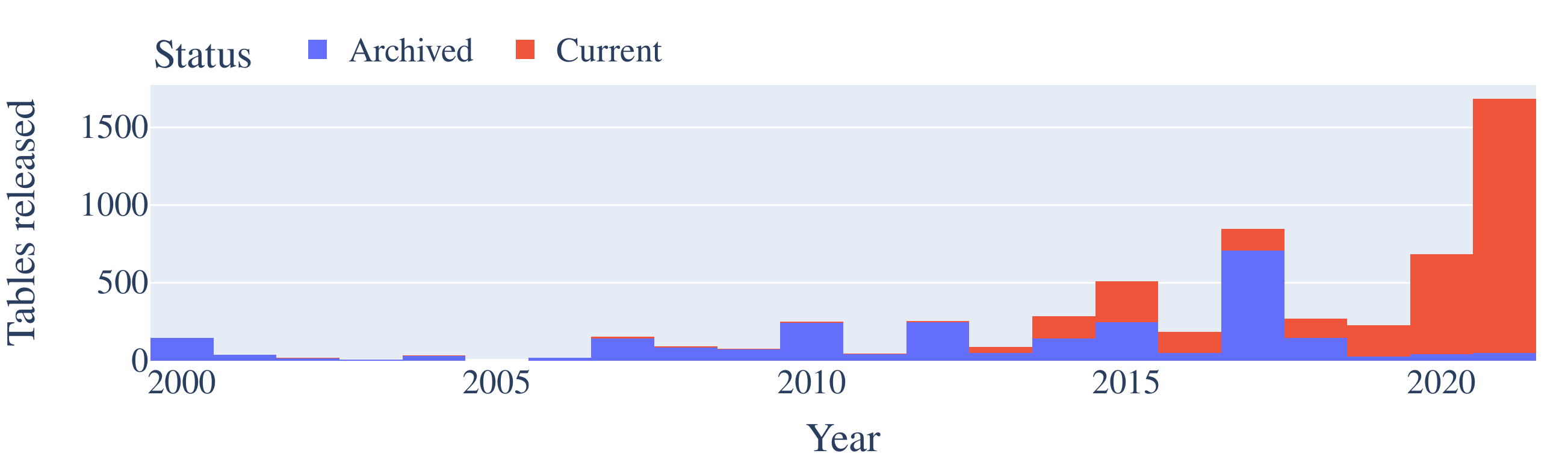}
    \caption{The release year of all tables available on \urlx{statcan.gc.ca}}
    \label{fig:table_release_history}
\end{figure*}

\begin{figure*}[h]
    \small
    \centering
    \includegraphics[width=\linewidth]{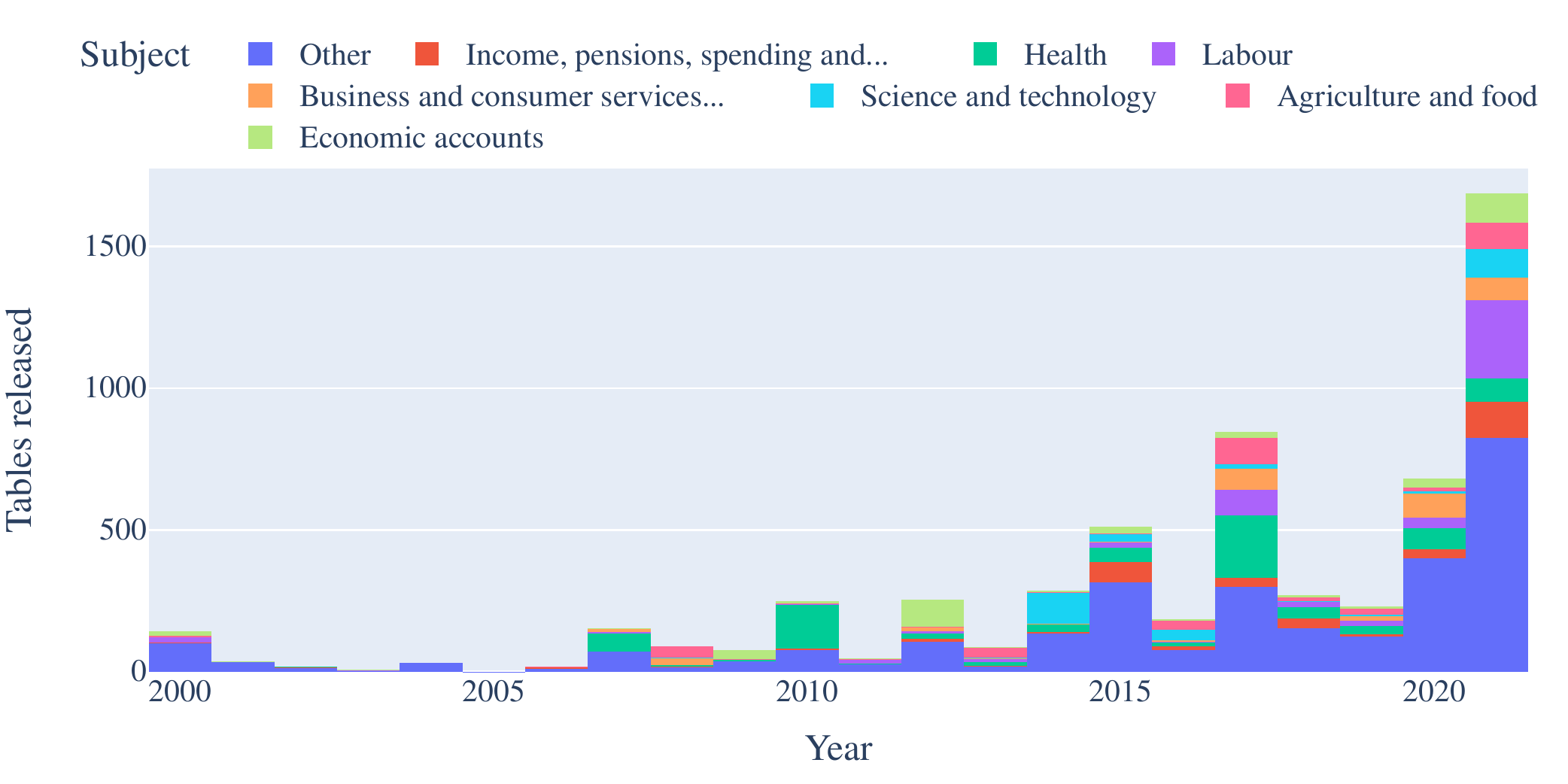}
    \caption{The release year of all tables by subject. Only the top 8 subjects are shown for readability.}
    \label{fig:table_release_history_by_subject}
\end{figure*}

Tables on Statcan will sometimes be updated regularly, whereas in other times they may only be released once. When they stop being updated and the information becomes outdated, they will be marked as archived. In some cases, a new version of an archive table may be created with substantial changes (such as new columns). As shown in \autoref{tab:release_statcan_table}, tables are released without a predetermined schedule, but has been on average increasing since 2000, with major peaks in 2017, which was caused by many health-related tables, and 2021, which was caused by the release of many labour, science, and income tables, as shown in \autoref{fig:table_release_history_by_subject}. Moreover,  \autoref{fig:table_release_history} shows that a majority of tables released between 2019 and 2021 are still up-to-date (current), whereas most of the tables before then have been archived.

\subsection{Supplementary Statistics}
\label{sec:appendix_supplementary_stats}

\subsubsection{Frequently requested tables}
\label{sec:appendix_frequently_requested_tables}
The most frequent tables are summarized in \autoref{tab:most_popular_tables}. Whereas 2 of them are sourced from the consumer price index survey (commonly used to track inflation), the 4 other tables cover more general and broad subjects like income, demography, business performance and crime. Each of those 4 tables are sourced from different surveys. One table is updated monthly, another semi-annually, and the rest are updated annually. The oldest table was updated in 2019, which is when the conversations started being recorded. 

\subsubsection{Table Frequency Statistics}

In tables \ref{tab:table_conversation_statistics} and \ref{tab:table_conversation_statistics_fr}, we can calculate that a table is returned on average 5.68 times (with standard deviation of 12.86) in English conversations and 4.25 times (standard deviation of 9.73) in French conversations. Thus, the most requested tables are disproportionately represented compared to less popular tables, and there's a very high variance in the number of time a table is used. In fact, there are 294 tables that appear only once in either splits (i.e., 28.9\%).

\subsubsection{Fine-grained Conversation Statistics}

In \autoref{tab:more_stats_conversation}, we observe that the number of messages and turns will vary significantly around the mean, with over 68\% conversations lasting between 2 and 7 turns. In extreme cases, a conversation can last up to 28 turns. Moreover, we also notice that, although most messages will have around 32 tokens, the longest message can have up to 1374 tokens; in those scenarios, we will see agents write a large body of text, and sometimes also copy and paste large amount of text (for example, from a database of templates) when responding to the user. Although those are usually sent in multiple consecutive messages within a turn, they may decide to send everything all at once.

\subsection{Implementation Details}
\label{sec:appendix_implementation_details}

This section provides the details for implementing the models in \autoref{sec:models}.

\paragraph{Implementing transformer models} All models based on the transformer architecture \citep{vaswani_attention_2017} were implemented using HuggingFace's library \citep{wolf_transformers_2020}. 

\paragraph{BM25} To facilitate reproducibility, we implemented the model in Gensim \citep{rehurek_software_2010}.

\paragraph{DPR and TAPAS-NQ} We used the base variant of DPR and the large variant of TAPAS-NQ. We use the DPR checkpoints that were trained on 320K questions from Natural Questions \citep{kwiatkowski_natural_2019} (NQ). During training, the networks were optimized with AdamW \citep{Loshchilov2017DecoupledWD} at a learning rate of $10^{-5}$ and zero weight decay. Based on the original work, the networks were trained for 30 epochs, with a batch size of 64 queries, positive passages, and hard negative passages (the latter are retrieved with BM25). Negative in-batch sampling was used to increase negative examples. To ensure reproducibility, the networks were trained on a single 32GB GPU and used gradient checkpointing \citep{chen_training_2016} to reduce memory usage. The conversation lengths was 512 tokens, and the metadata token lengths were 128 for title, 256 for basic information (defined in \autoref{sec:table_specs}), and 512 for the rest. 

\paragraph{T5} We used an Adafactor optimizer \citep{pmlr-v80-shazeer18a} with a learning rate of 0.001. We used batch sizes of 16 with 8 steps of gradient accumulation and gradient checkpointing to reproduce the batch size of 128 samples in the original implementation. The models were trained on a single 32GB GPU for 10 epochs. The source and target lengths were respectively 512 and 256 tokens, where the source was truncated from the right to ensure that the latest messages remained after truncation. We used a beam size of 4 and length penalty of 0.6 following the original implementation.

\paragraph{Training time} For the English split, each variant of DPR can be trained in 68 minutes on a A100 GPU. Each large variant of TAPAS-NQ takes 15h to train on a V100 GPU. Each variant of T5 can be trained in 16h on a V100 GPU. All DPR results can be reproduced in 8h, TAPAS in 90h, and T5 in 96h, and proportionally less time would be needed for the French split.

\subsection{Modeling the French subsets}
\label{sec:french_modeling}

\begin{figure*}[h]
    \small
    \centering
    \includegraphics[width=0.7\linewidth]{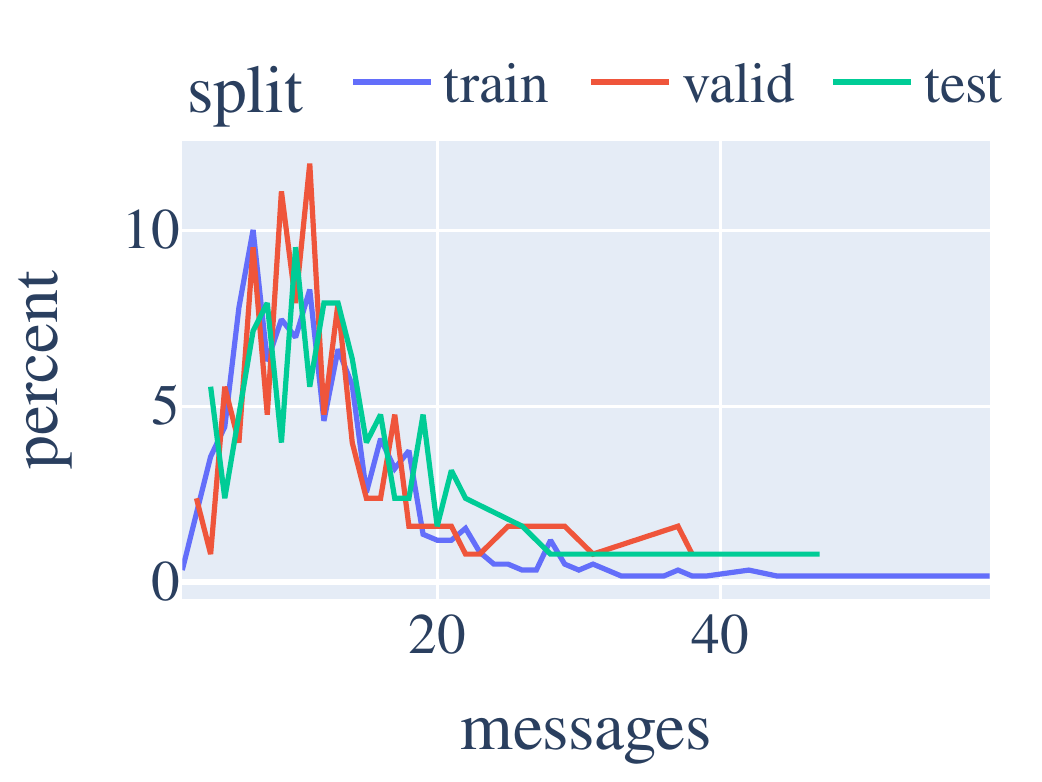}
    \caption{Histogram of messages by conversation in the French task splits. English split in \autoref{fig:turn_msg_histogram}.}
    \label{fig:turn_msg_histogram_fr}
\end{figure*}

\paragraph{Basic statistics} The number of messages by conversation varies between 2 and 59 with a median of 11 for the English split (see \autoref{fig:turn_msg_histogram_fr} for the distribution). Based on \autoref{tab:table_conversation_statistics_fr}, there's on average 3.9 turns but 12.3 messages. On average, there are over 30 tokens for each message (using the T5 tokenizer). 

\paragraph{Language splitting} In order to determine the language of each conversation, we used two popular language identification libraries: \textit{langid.py} \citep{lui_cross-domain_2011} and a fasttext network finetuned for language detection \citep{joulin-etal-2017-bag}. After apply the models on every conversation, we only retain the conversations with matching language labels (both English or both French). 

\paragraph{Training and evaluation} The training procedure and evaluation on the French subsets follow exactly the tasks specified in \autoref{sec:tasks}. 

\paragraph{Modeling response generation} Instead of T5, we used the multilingual T5 model by \citet{xue_mt5_2021} as it naturally handles text in French. 

\paragraph{Modeling retrieval} We used a variant of DPR derived from CamemBERT \cite{martin-etal-2020-camembert} and trained on three French Q\&A datasets \citep{keraron-etal-2020-project, dhoffschmidt_fquad_2020, kabbadj_french-squad_2021} by \citet{etalab_lab-ia_dpr_2021}.

\paragraph{Retrieval results} In \autoref{tab:full_retrieval_results_french}, we observe that, unlike the English split, adding member items to the basic information or to the title improves validation results but not test results, which likely indicates overfitting. However, we notice a high variance between the runs, which makes it difficult to determine whether member items is helpful. Both overfitting and high variance are likely caused by the smaller size of the training set. Moreover, BM25 perform extremely poorly on any metadata view, which can also be linked to the dataset size.

\paragraph{Generation results} In \autoref{tab:full_generation_results}, we notice a significant decrease across all metrics, with the title accuracy being consistently 0\%. This is likely because the French split is significantly smaller, yet remains as complex as the English split, which becomes challenging for mT5 to model. In the case of title accuracy, we found 55 instances in the French test split where the title is in the target text (i.e., returned by an agent). However, in 54 cases, the augmented mT5 returned a generic reply (e.g., ``Veuillez patienter pendant que j'effectue une recherche.") instead of the expected title, which indicates that mT5 is incapable of determining when it is relevant to return a title and can't generate non-templated responses.

\subsection{Responses generated by T5}
\label{sec:appendix_case_study_t5}

In this section, we select a few conversations from the validation set and examine the messages generated by T5 and T5 augmented with DPR-retrieved titles (T5+D).

\paragraph{Common and uncommon responses} In \autoref{tab:sample_conversation_4890}, we notice that both T5 and T5+D are capable of generating common speech acts like ``Thank you'' and ``Please wait...'', but struggles when faced with an unfamiliar situation (having to ask for clarification for a user that has been accidentally disconnected).

\paragraph{Common table} In \autoref{tab:sample_conversation_4890_part_2}, among the retrieved tables ($R_i$) titles, the first one was partially correct. Both T5 and T5+D extended the title and also output the desired ID, matching the expected agent's response, which is one of the most popular table in the training set (see \autoref{tab:most_popular_tables}).

\paragraph{Multiple tables, date selection} In \autoref{tab:sample_conversation_8500}, we notice that T5 only returns one of the two tables that the agent returned. On the other hand, the correct tables were retrieved by DPR, but T5+D failed to select the ones with the correct dates (it selected June 2019 instead of December 2019) but the select were otherwise relevant. 

\paragraph{Verbosity of explanations} T5+D additionally provided a paragraph of explanation while linking to relevant resources, both on the StatCan website (non-tabular) and external resources. This is because T5+D memorized this information during training, and simply replaced the tables with the updated dates (\autoref{tab:sample_conversation_1628}).

\paragraph{Uncommon tables} In the conversations shown in Tables \ref{tab:sample_conversation_8960}, \ref{tab:sample_conversation_21533}, \ref{tab:sample_conversation_19568}, the retrieved table appears 10 times in the training set, which is significantly less common than the table retrieved in \autoref{tab:sample_conversation_4890_part_2}. For \autoref{tab:sample_conversation_8960}, the table returned by the agent was not retrieved by DPR, leading to T5+D returning the first title retrieved. However, in \autoref{tab:sample_conversation_21533}, the correct title was retrieved by DPR (title \#4), yet T5+D failed to use that correct title in the generated message. As for \autoref{tab:sample_conversation_19568}, the agent gave a hint by stating ``As a standard product, we have tables about employment by industry'', which was correctly acknowledged by DPR as the second retrieved title perfectly matches the PID of the table in the agent's response. However, T5+D fails again at selecting the correct title, instead opting to return a generic response (``Please hold while I find the information''), and T5 hallucinates a PID that is different from the title it generated (both of which are wrong).

\paragraph{Tables unseen during training} Among the tables that do not appear in the training set (see \autoref{tab:table_differences_and_overlaps} for more information), there are seven that appear 3 or more times in the validation or test sets (\autoref{tab:tables_not_in_train_with_three_appearances}). In \autoref{tab:sample_conversation_15376}, we see a conversation where DPR retrieves the correct title, which is correctly returned by T5+D, whereas T5 fails to return it. On the other hand, when DPR also correctly retrieves the title in \autoref{tab:sample_conversation_11869}, T5+D fails to return it, as it was likely mislead by the agent saying ``Unfortunately,...''.

\subsection{Statistics Tests}
\label{sec:appendix_stats_tests}

To back the claims in \autoref{sec:results_analysis}, we performed multiple single-tailed Welch t-tests, using the mean and corrected standard deviation from \ref{tab:full_retrieval_results_english}. The null hypotheses are that means of experiments A are different from the means of experiments B, across 3 runs. Unless otherwise specified, we use R@1 on the test split.

\begin{enumerate}
    \item Claim: Adding basic information and member items to title results in a significant difference for DPR. With $A$ being the model using only title, and $B$ using basic + member, our p-value is 0.014.
    \item Claim: For DPR, using member item result in drastic decrease. With $A$ being the model using only member and $B$ using title + member, our p-value is 0.00086.
    \item Claim: TAPAS-NQ performs better with title and member items compared to the full table. With $A$ being the model using the full table, and $B$ using title + member, our p-value is 0.016.
    \item Claim: In \autoref{fig:valid_vs_test_acc}, the validation recall@10 are higher than the test split for TAPAS, DPR Title and DPR Basic + member. With $A$ being the validation score and $B$ the test scores, the p-values are respectively $0.00197, 2.18 \times 10^{-5}, 0.00014$.
\end{enumerate}

\newpage

\section{Dataset Card}
\label{sec:appendix_dataset_card}

This section presents a dataset card that follows the format proposed by \citet{lhoest2021datasets}, which was inspired by \citet{mitchell_model_2019} and \citet{Gebru2021DatasheetsFD}.

\paragraph{Summary} The StatCan Dialogue Dataset consists of over 20K+ conversations between agents working at Statistics Canada (StatCan) and users who are visiting StatCan's website and need support via the official live chat system. 

\paragraph{Tasks} A subset of 19K conversation turns is used to build two tasks:

\begin{enumerate}
    \item Automatic retrieval of relevant tables based on a on-going conversation. For each partial conversation, the task is to return the ID of the most likely table returned by an agent. This is evaluated using the recall@k metric.
    \item Automatic generation of appropriate agent responses at each turn. For each partial conversation, the task is to return the most likely response by an agent, including link to a relevant table. This is evaluated using four metrics described in \autoref{sec:response_generation_task}.
\end{enumerate}

\paragraph{Leaderboard} The leaderboard and submission instructions can be found on the project webpage. Each submission will be accompanied with a tag indicating if:
\begin{itemize}
    \item It was self-reported;
    \item The submissions were externally evaluated;
    \item The inference was reproduced following provided instructions;
    \item The complete training process was independently reproduced.
\end{itemize}

\paragraph{Languages} The conversations were held in Canadian English (en-CA) and Canadian French (fr-CA).

\subsection{Dataset Structure}

\subsubsection{Data Instances}

\paragraph{Conversation} A full example of a conversation can be found in \autoref{tab:full_example_conversation}. Instances for each user intent can be found in \autoref{tab:examples_user_intents}, and two conversations with annotated dialogue acts can be found in \autoref{tab:sample_linguistic_analysis}. For our case study in \ref{sec:appendix_case_study_t5}, we show partial conversations in Tables \ref{tab:sample_conversation_4890_part_2}, \ref{tab:sample_conversation_8960}, \ref{tab:sample_conversation_21533}, \ref{tab:sample_conversation_1628}, \ref{tab:sample_conversation_8500}, \ref{tab:sample_conversation_19568}, \ref{tab:sample_conversation_15376}, \ref{tab:sample_conversation_11869}.

\paragraph{Tables} The complete metadata of a table can be found in \autoref{tab:sample_full_info}, which can be access at \href{https://doi.org/10.25318/3210035401-eng}{doi.org/10.25318/3210035401-eng}. \autoref{tab:most_popular_tables} shows the basic information for the most popular tables.

\subsubsection{Data Fields}

\paragraph{Full dataset} A CSV file with the following fields is provided:
\begin{itemize}
    \item \verb|conversation|: The partial conversation (before a table is returned) in JSON format.
    \item \verb|conversation_index|: A unique index that serves at identifying the conversation outside of this task.
    \item \verb|conversation_processed|: The conversation converted into a readable text format, with extra information (such as timestamp) removed, the URLs replaced with a special tag, and separation tags (\verb|</s>|) added.
\end{itemize}

\paragraph{Retrieval task} CSV files with the following fields is provided for each split:

\begin{itemize}
    \item \verb|conversation|
    \item \verb|conversation_index|
    \item \verb|conversation_processed|
    \item \verb|target_pid|: The product ID of the table that is returned by the agent
    \item \verb|language|: The language reported by the live chat system, which may not always be accurate due to mislabeling.
    \item \verb|ft_detected_lang|: The language predicted by fastText.
    \item \verb|ft_detected_lang|: The score output by fastText.
    \item \verb|lid_detected_lang|: The language predicted by langid.py.
    \item \verb|lid_detected_prob|: The score output by langid.py.
\end{itemize}

\paragraph{Metadata} The metadata that was used during retrieval is provided as a CSV file with the following fields (one for each of the 5907 tables):
\begin{itemize}
    \item \verb|pid|: The product ID of the table
    \item \verb|title|: The title of the table
    \item \verb|basic_info|: The basic information in the textual format
    \item \verb|member_info|: The member items enumerated as text
    \item \verb|footnote_info|: The footnotes enumerated as text
    \item \verb|full_info|: The basic information, member items and footnotes all in a single entry
    \item \verb|x_and_y|: An combination of two items above, for example \verb|x=title| and \verb|y=footnotes|.
    \item \verb|*_fr|: All of the above are also available in French, indicated by the suffix \verb|_fr|.
    
\end{itemize}

\paragraph{Generation task} CSV files with the following fields is provided for each split:
\begin{itemize}
    \item \verb|source|: Equivalent to \verb|conversation|.
    \item \verb|source_processed|: Equivalent to \verb|conversation_processed|.
    \item \verb|target|: The message written by the agent following the conversation.
    \item \verb|target_processed|: The message written by the agent following the conversation, with URLs replaced with a special tag.
    \item \verb|conversation_index|
\end{itemize}

An augmented variant of each CSV file for the conversation task is provided with all of the above as well as the following fields:
\begin{itemize}
    \item \verb|source_augmented|: The same content as \verb|conversation|, appended with the title of the top-5 tables retrieved by best DPR variant trained on the basic information.
    \item \verb|target_augmented_1|: The same content as \verb|source_augmented|, but using only the first table instead of top-5.
\end{itemize}

\paragraph{Conversation JSON formatting}
A conversation follows the following JSON format:

\begin{small}
\begin{verbatim}
[
  ...
 {'timestamp': '13.03.2019 17:03:22',
  'actor': 'user',
  'actor_name': '<NAME>',
  'text': "I'll take a look at that",
  'urls': []},
 {'timestamp': '13.03.2019 17:04:12',
  'actor': 'operator',
  'actor_name': 'Kelly C',
  'text': 'Building permits...',
  'urls': ['https:...']},
  ...
]
\end{verbatim}
\end{small}

\subsubsection{Data Split}

The retrieval splits has the following number of samples:
\begin{itemize}
    \item Train: 3782 (en); 869 (fr)
    \item Validation: 799 (en); 201 (fr)
    \item Test: 870 (en); 141 (fr)
\end{itemize}

They correspond to the number of queries in \autoref{tab:table_conversation_statistics} and \autoref{tab:table_conversation_statistics_fr} because each query results in a table being retrieved, which can happen more than once in a conversation. In such cases, the partial conversations will be truncated at different turns in the conversation.

The generation splits have the following number of samples:
\begin{itemize}
    \item Train: 21582 (en); 3977 (fr)
    \item Validation: 4464 (en); 861 (fr)
    \item Test: 4850 (en); 884 (fr)
\end{itemize}

There are fewer samples than the number of messages in  \autoref{tab:table_conversation_statistics} and \autoref{tab:table_conversation_statistics_fr} because the goal of the task is only to predict the messages that will be written by the agent. Just like the retrieval task, the partial conversations will be truncated at different turns in the conversation.

\subsection{Dataset Creation}

\subsubsection{Curation Rationale}

\autoref{sec:introduction} extensively motivates the curation of the dataset. To summarize, we enumerate the major points:

\begin{itemize}
    \item Data from real users: We wanted a dataset that captures the linguistic challenges that exist in the real world
    \item Task-oriented dialogue: We wanted tasks with the specific goal of helping live chat users in their search of statistics.
    \item Real-world applications: Our model can be directly applied other statistics offices that want to set up a chat system, and our dataset will be useful for any organization that has (1) a chat system, and (2) a database of tables.
    \item Multilingual dialogue: We wanted to build models that can handle languages beyond English, which is why we also offer French versions of our tasks.
\end{itemize}

\subsubsection{Source Data}

\paragraph{Conversations} The data was retrieved from the live chat system on \urlx{statcan.gc.ca}, which was anonymized by the development team at StatCan. The conversations happened between March 2019 and March 2021.

\paragraph{Tables} The tables are publicly available and were downloaded following the instructions in the Web Data Service user guide: \href{https://www.statcan.gc.ca/en/developers/wds/user-guide}{statcan.gc.ca/en/developers/wds/user-guide}. The tables were released by Statistics Canada between 2000 and 2021. The data are either collected directly by Statistics Canada (e.g., through a census or a survey) or were compiled from existing sources (such as private sector organizations and government agencies) into official statistics. Existing sources include:

\begin{itemize}
    \item Administrative data: Collected by government or the private sector as part of ongoing operations, and include records of birth and death, taxes, border control, and satellite data. 
    \item Microdata Linkage: Existing information is linked to create new data. The existing information may not always be available publicly (for privacy purposes), thus linkage could add new information that was previously unavailable, while protecting the confidentiality of the public.
    \item Open data: Machine-readable and freely available data sourced from various channels (e.g., OpenStreetMap).
    \item Web scraping: Data from the internet that were scraped by Statistics Canada (this excludes personal information and ``any information that will not be used to produce statistical output").
\end{itemize}

\subsubsection{Annotations}

The dataset does not contain any additional annotations beyond the ones collected through the live chat system and included in the table metadata.

\subsubsection{Personal and Sensitive Information}

Personal and sensitive information were removed programmatically from the conversations, and officially published tables only contain aggregated information that preserve the confidentiality of the participants. Although the removal process is highly advanced, there is a non-zero chance that some information can be used to reconstruct the profile of a user. For this reason, the access to the data will require researchers to sign-up and agree to the terms of use, and any derivative must be shared on the same platform and include the same terms.

\subsection{Considerations for Using the Data}

\subsubsection{Social Impact of Dataset}

The purpose of this dataset is develop and evaluate models that can assist knowledge workers in finding relevant tables from a data source. By providing a specialized retrieval system capable of returning more relevant results compared to general purpose, the productivity of the knowledge workers can be increased. For public agencies and statistics offices, this would benefit many live chat users interested in statistics related to a certain community.

\subsubsection{Discussion of Biases}

As discussed in \autoref{sec:limitations}, there are always risks of toxicity in online discourses, which means that the live users may exhibit negative biases in their messages. However, the StatCan agents are tasked to communicate with online users in a professional manner. Thus, researchers should not use this dataset to build models that generate messages written by live users, and any model trained on the dataset should not be used in scenarios where biases can negatively impact stakeholders.

\subsubsection{Other Known Limitations}

\autoref{sec:limitations} describes other known limitations.

\subsection{ Additional Information }

\subsubsection{ Dataset Curators }

The dataset was curated by the authors of this paper based on the original data collected and processed by StatCan developers and agents.

\subsubsection{Licensing Information }

The conversations use a custom license, which needs to be accepted by researchers interested in accessing the conversation. The tables are released under the the Statistics Canada Open Licence: \href{https://www.statcan.gc.ca/en/reference/licence}{statcan.gc.ca/en/reference/licence}.

\begin{table}[t]
    \small
    \centering
    \begin{subtable}[]{\linewidth}
        \centering
        \begin{tabular}{lrrrr}
        \toprule
        {} &  Train &  Valid &  Test &  Overall \\
        \midrule
        Train   &      0 &    470 &   535 &        0 \\
        Valid   &     41 &      0 &   199 &        0 \\
        Test    &    145 &    238 &     0 &        0 \\
        Overall &    181 &    610 &   571 &        0 \\
        \bottomrule
        \end{tabular}
        \caption{Differences}
        \label{tab:splits_table_diff}
    \end{subtable}
    \hfill
    \begin{subtable}[]{\linewidth}
        \centering
        \begin{tabular}{lrrrr}
        \toprule
        {} &  Train &  Valid &  Test &  Overall \\
        \midrule
        Train   &    778 &    308 &   243 &      778 \\
        Valid   &    308 &    349 &   150 &      349 \\
        Test    &    243 &    150 &   388 &      388 \\
        Overall &    778 &    349 &   388 &      959 \\
        \bottomrule
        \end{tabular}
        \caption{Overlaps}
        \label{tab:splits_table_intersection}
    \end{subtable}
    \hfill
    \begin{subtable}[]{\linewidth}
        \centering
        \begin{tabular}{lrrrr}
        \toprule
        {} &  Train &  Valid &  Test &  Overall \\
        \midrule
        Train   &      0 &    154 &   184 &        0 \\
        Valid   &     18 &      0 &    64 &        0 \\
        Test    &     44 &     60 &     0 &        0 \\
        Overall &     61 &    197 &   201 &        0 \\
        \bottomrule
        \end{tabular}
        \caption{Differences (French)}
    \end{subtable}
    \hfill
    \begin{subtable}[]{\linewidth}
        \centering
\begin{tabular}{lrrrr}
\toprule
{} &  Train &  Valid &  Test &  Overall \\
\midrule
Train   &    224 &     70 &    40 &      224 \\
Valid   &     70 &     88 &    24 &       88 \\
Test    &     40 &     24 &    84 &       84 \\
Overall &    224 &     88 &    84 &      285 \\
\bottomrule
\end{tabular}
        \caption{Overlaps (French)}
    \end{subtable}
    \caption{Number of tables (a,c) differing and (b,d) overlapping between each split (subset used for both tasks). The difference is computed as row - column. Summarized results in \autoref{tab:table_conversation_statistics}.}
    \label{tab:table_differences_and_overlaps}
\end{table}

\begin{table}[h]
    \small
    \centering
\begin{tabular}{lrrrr}
\toprule
          Dataset &  Train &  Valid &  Test &   All \\
\midrule
          \# Conv. &    562 &    122 &   109 &   793 \\
          \# Turns &   2147 &    472 &   439 &  3058 \\
       \# Messages &   6807 &   1492 &  1389 &  9688 \\
        \# Queries &    869 &    201 &    141 &   1211 \\
   \# Tokens / Msg &  47.25 &  49.86 & 40.14 & 46.63 \\
  \# Turns / Conv. &   3.82 &   3.87 &  4.03 &  3.86 \\
    \# Msg / Conv. &  12.11 &  12.23 & 12.74 & 12.22 \\
\# Queries / Conv. &   1.55 &   1.65 &   1.31 &   1.53 \\
     \midrule
         \# Tables &    224 &     88 &    84 &   285 \\
     \# New tables &      0 &     18 &    44 &    61 \\
   \# Dims / Table &    3.4 &    3.5 &   3.5 &   3.4 \\
   \# Mbrs / Table &  205.8 &  249.9 & 164.1 &   190 \\
  \# Notes / Table &   20.3 &   26.3 &  23.4 &    21 \\
\bottomrule
\end{tabular}
    \caption{Statistics for the retrieval and generation tasks for the French split. See \autoref{tab:table_conversation_statistics} for the English split.}
    \label{tab:table_conversation_statistics_fr}
\end{table}

\begin{table}[h]
    \small
    \centering
\begin{tabular}{lrrrr}
\toprule
          Dataset &  Train &  Valid &  Test &   All \\
\midrule
 Max Tokens / Msg &   1243 &    857 &   912 &  1243 \\
Max Turns / Conv. &     28 &     18 &    21 &    28 \\
  Max Msg / Conv. &     78 &     58 &    73 &    78 \\
   Max Qs / Conv. &     15 &     11 &    11 &    15 \\
 Std Tokens / Msg &      1 &      1 &     1 &     1 \\
Std Turns / Conv. &   2.58 &   2.36 &  2.62 &  2.56 \\
  Std Msg / Conv. &   8.53 &   7.55 &  8.98 &  8.47 \\
   Std Qs / Conv. &   1.27 &   1.24 &  1.18 &  1.25 \\
   \midrule
 Max Tokens / Msg &   1374 &   1373 &   352 &  1374 \\
Max Turns / Conv. &     15 &     10 &    15 &    15 \\
  Max Msg / Conv. &     59 &     38 &    47 &    59 \\
   Max Qs / Conv. &     15 &     15 &     5 &    15 \\
 Std Tokens / Msg &      1 &      2 &     2 &     1 \\
Std Turns / Conv. &   2.21 &   2.08 &   2.1 &  2.17 \\
  Std Msg / Conv. &   7.02 &   6.74 &  6.61 &  6.91 \\
   Std Qs / Conv. &   1.84 &   2.18 &  0.77 &  1.79 \\
\bottomrule
\end{tabular}
    \caption{More statistics (max and standard deviation) at the conversation level (top: English, bottom: French), following \autoref{tab:table_conversation_statistics}.}
    \label{tab:more_stats_conversation}
\end{table}

\begin{table}[h]
    \small
    \centering
\begin{tabular}{lrr}
\toprule
Year Released & \# Archived &  \# Up-to-date \\
\midrule
2000 &       145 &        0 \\
2001 &        39 &        0 \\
2002 &        14 &        5 \\
2003 &         8 &        0 \\
2004 &        31 &        2 \\
2005 &         1 &        0 \\
2006 &        18 &        0 \\
2007 &       142 &       11 \\
2008 &        85 &        7 \\
2009 &        72 &        5 \\
2010 &       243 &        8 \\
2011 &        45 &        3 \\
2012 &       248 &        8 \\
2013 &        49 &       40 \\
2014 &       142 &      145 \\
2015 &       247 &      264 \\
2016 &        52 &      134 \\
2017 &       710 &      136 \\
2018 &       148 &      122 \\
2019 &        27 &      203 \\
2020 &        43 &      640 \\
2021 &        52 &     1633 \\
\bottomrule
\end{tabular}
    \caption{Release of Statcan tables over the years}
    \label{tab:release_statcan_table}
\end{table}

\begin{table*}[h]
    \small
    \centering
    \begin{tabular}{l c p{0.65\linewidth}}
        \toprule
\textbf{Merged Acts}     & \multicolumn{1}{l}{\textbf{Count}} & \textbf{Original Speech Acts}                                                     \\ \midrule
answer                & 50                                 & answer, confirm, disconfirm                                                       \\
request               & 31                                 & request, acceptOffer, declineOffer                                                \\
timeManagement        & 28                                 & stalling, pausing                                                                 \\
inform                & 63                                 & inform, agreement, disagreement, correction                                       \\
infoSeekingQuestions  & 25                                 & propositionalQuestion, checkQuestion, setQuestion, choiceQuestion, directQuestion \\
promise               & 18                                 & promise, acceptRequest, declineRequest                                            \\
autoFeedback          & 25                                 & positiveAutoFeedback, negativeAutoFeedback                                        \\
offer                 & 11                                 & offer                                                                             \\
instruct              & 18                                 & instruct                                                                          \\
clarificationQuestion & 16                                 & clarificationQuestion                                                             \\
greeting              & 28                                 & initialGreeting, returnGreeting                                                   \\
selfIntroduction      & 17                                 & initialSelfIntroduction, returnSelfIntroduction                                  \\
thanking              & 47                                 & thanking                                                                          \\
acceptThanking        & 15                                 & acceptThanking                       
\\
        \bottomrule
    \end{tabular}
    \caption{Merged and original speech acts occurring in 100 turns in conversations from the validation set. This table is summarized in \autoref{tab:merged_speech_act_counts}.}
    \label{tab:iso_speech_acts_count}
\end{table*}

\begin{table*}[h]
    \small
    \centering
\begin{tabular}{llllllll}
\toprule
     &                & \multicolumn{3}{l}{Valid} & \multicolumn{3}{l}{Test} \\
Model & Encoded &          R@1 &         R@10 &         R@20 &          R@1 &         R@10 &         R@20 \\
\midrule
DPR & Basic &  37.8 ± 1.82 &  73.1 ± 1.02 &  80.8 ± 0.51 &  14.7 ± 1.13 &  45.0 ± 0.24 &  55.0 ± 1.67 \\
     & Basic + footnote &  40.0 ± 0.64 &  73.3 ± 0.85 &  81.3 ± 1.19 &  13.9 ± 0.37 &  44.4 ± 2.73 &  54.2 ± 2.65 \\
     & Basic + member &  40.0 ± 0.44 &  74.5 ± 1.38 &  81.6 ± 0.55 &   15.7 ± 0.7 &  46.2 ± 0.44 &  56.3 ± 0.58 \\
     & Member &  28.0 ± 0.47 &  62.1 ± 1.07 &  71.4 ± 1.06 &  10.7 ± 0.88 &  35.0 ± 0.57 &  46.3 ± 0.83 \\
     & Title &  37.8 ± 0.21 &  70.9 ± 0.67 &  78.2 ± 1.18 &  13.9 ± 0.59 &  43.8 ± 1.22 &  53.4 ± 1.42 \\
     & Title + footnote &  39.5 ± 0.69 &  73.2 ± 0.81 &  80.8 ± 0.47 &   13.3 ± 0.7 &  43.6 ± 0.46 &   53.6 ± 1.5 \\
     & Title + member &  38.6 ± 1.19 &  72.1 ± 0.92 &  79.4 ± 0.72 &  14.3 ± 0.59 &  45.1 ± 0.44 &  54.2 ± 1.07 \\
\midrule
TAPAS & Table &  16.9 ± 4.26 &  45.9 ± 5.46 &  55.5 ± 3.13 &   3.9 ± 0.65 &  17.2 ± 2.83 &  24.6 ± 3.45 \\
     & Title &  21.5 ± 0.33 &  54.7 ± 3.39 &  63.7 ± 1.38 &   4.7 ± 0.53 &  20.2 ± 0.85 &  28.7 ± 1.98 \\
     & Title + member &   24.6 ± 3.6 &  54.9 ± 3.34 &  63.7 ± 2.74 &   6.1 ± 0.98 &  22.1 ± 2.66 &  31.5 ± 3.02 \\
TAPAS-NQ & Table &  25.3 ± 3.46 &  60.0 ± 2.43 &  70.0 ± 2.89 &    5.3 ± 0.8 &  22.8 ± 2.32 &  32.0 ± 2.48 \\
     & Title &  26.5 ± 1.85 &  64.4 ± 1.95 &  71.6 ± 2.32 &   6.3 ± 0.64 &  27.4 ± 1.84 &  36.1 ± 2.53 \\
     & Title + member &  29.8 ± 2.74 &  62.9 ± 1.35 &  71.5 ± 0.63 &   7.4 ± 0.73 &  30.0 ± 1.46 &  39.3 ± 0.86 \\
\midrule
BM25 & Basic &          1.0 &          7.5 &         10.3 &          1.0 &          7.2 &          9.8 \\
     & Basic + footnote &          0.9 &          2.4 &          3.1 &          0.9 &          2.8 &          3.9 \\
     & Basic + member &          0.1 &          1.3 &          2.6 &          0.2 &          2.4 &          3.8 \\
     & Full &          0.3 &          1.3 &          1.9 &          0.2 &          1.1 &          2.0 \\
     & Title &          2.0 &         10.3 &         13.8 &          2.5 &         10.9 &         14.0 \\
     & Title + footnote &          0.8 &          2.3 &          3.1 &          0.7 &          2.9 &          4.0 \\
     & Title + member &          0.0 &          1.4 &          2.4 &          0.3 &          2.3 &          3.8 \\
\bottomrule
\end{tabular}
    \caption{Full retrieval results for the English splits. The values reported are in recall \% at $k$. DPR and TAPAS were run 3 times and averaged (standard deviation given after ±). Selected results in \autoref{tab:selected_retrieval_results_english}.}
    \label{tab:full_retrieval_results_english}
\end{table*}

\begin{table*}[h]
    \small
    \centering
\begin{tabular}{llllllll}
\toprule
     &                & \multicolumn{3}{l}{Valid} & \multicolumn{3}{l}{Test} \\
Model & Encoded &          R@1 &         R@10 &         R@20 &          R@1 &         R@10 &         R@20 \\
\midrule
DPR & Basic &  19.7 ± 1.52 &  57.4 ± 2.55 &  67.0 ± 5.01 &  12.5 ± 1.78 &  40.0 ± 1.08 &  48.5 ± 4.27 \\
     & Basic + footnote &   23.2 ± 3.8 &  55.4 ± 1.88 &  66.8 ± 1.04 &  13.9 ± 1.48 &  35.9 ± 0.82 &  49.9 ± 3.91 \\
     & Basic + member &  22.7 ± 4.35 &  57.9 ± 2.45 &  67.7 ± 3.48 &  12.8 ± 2.13 &  38.1 ± 2.17 &  47.0 ± 1.78 \\
     & Title &  23.7 ± 2.92 &  58.5 ± 2.24 &  69.8 ± 2.35 &  14.7 ± 2.28 &   35.7 ± 3.2 &  48.2 ± 2.84 \\
     & Title + footnote &  23.2 ± 2.74 &  54.7 ± 3.59 &  64.7 ± 2.28 &  14.4 ± 1.48 &  36.6 ± 2.69 &  45.4 ± 2.13 \\
     & Title + member &  24.2 ± 1.52 &  55.4 ± 4.51 &  65.3 ± 2.01 &  14.4 ± 2.28 &  39.5 ± 0.82 &  48.0 ± 2.05 \\
\midrule
BM25 & Basic &          0.0 &          0.0 &          0.0 &          0.0 &          0.0 &          0.0 \\
     & Basic + footnote &          0.0 &          0.0 &          0.0 &          0.0 &          0.0 &          0.0 \\
     & Basic + member &          0.5 &          1.0 &          1.0 &          0.0 &          0.7 &          0.7 \\
     & Full &          0.0 &          0.0 &          0.5 &          0.0 &          0.0 &          0.0 \\
     & Title &          0.0 &          0.5 &          0.5 &          0.0 &          0.0 &          0.7 \\
     & Title + footnote &          0.0 &          0.0 &          0.0 &          0.0 &          0.0 &          0.0 \\
     & Title + member &          0.0 &          0.5 &          1.0 &          0.0 &          0.7 &          0.7 \\
\bottomrule
\end{tabular}
    \caption{Full retrieval results for the French splits. The values reported are in recall \% at $k \in \{1,10,20\}$. DPR was run 3 times and averaged (standard deviation given after ±).}
    \label{tab:full_retrieval_results_french}
\end{table*}

\begin{table*}[h]
    \small
    \centering
\begin{tabular}{l p{0.90\linewidth}}
\toprule
   & \textbf{Obtaining meta-information of table}\\
U: &  Hi I have a question on the date of new census \\
A: &  Hello my name is Graham, how may I help you? \\
U: &  Hi Graham \\
U: &  I am currently looking at : Census of Canada and National Household Survey \\
U: &  I see that the last census were made in 2011 and 2016 \\
U: &  should I expect a new census in the coming month in 2020 ? \\
A: &  no - the next Census is going to be taking place in May 2021 \\
A: &  it's every 5 years \\
U: &  Ok perfect thank you \\
A: &  you're welcome! have a good day \\
\midrule
& \textbf{Help with user interface}\\
U: & Hello Graham - is there way I can see total annual values instead of monthly?\\
U: & for - \href{https://www150.statcan.gc.ca/t1/tbl1/en/cv!recreate.action?pid=3410017501&selectedNodeIds=1D1,2D13,2D14,2D19,2D28,4D1&checkedLevels=2D1&refPeriods=20200201,20200601&dimensionLayouts=layout3,layout3,layout3,layout2,layout2&vectorDisplay=false}{https://www150.statcan.gc.ca/t1/tbl1/en/cv!recreate.action?...}\\
A: & please hold - I will see if a table like this with an annual frequency is available\\
U: & wonderful thank you\\
A: & unfortunately no, the only active table on construction investment is the monthly one I sent you - there is an inactive quarterly table Investment in Building Construction (https://www150.statcan.gc.ca/n1/en/surveys/5014)\\
A: & it is also possible to download the monthly data in .CSV form, add together the months in a spreadsheet to get annual totals\\
A: & or if you'd prefer, you can make a custom request to our analysts (a fee may apply).Please send us your request, clearly indicating the variables, geographic areas, time frame and data frequency (i.e., annually, monthly or quarterly) needed at infostats@canada.ca (mailto:infostats@canada.ca)  and refer to Case number: 986005 so we can access the transcript of this live chat.               Your request must include your name, address, postal code, and phone number (please do not share your personal information on the live chat.)An agent will get in touch with you to provide an estimate and proceed with the payment, if applicable.You may also submit your request by telephone at 1-800-263-1136.\\
U: & thats fine! thank you very much :)\\
U: & very helpful\\
U: & :)!\\
U: & have a good day!\\
A: & glad I could help! you too\\
\midrule
& \textbf{Help with user interface} \\
U: &  Hello, Do you know where I can download the dataset for General Social Survey Cycle 32? \\
A: &  Please wait while I try to find this information for you. \\
U: &  Thank you \\
U: &  (I'm referring to this one: \href{https://www23.statcan.gc.ca/imdb/p2SV.pl?Function=getSurvey&SDDS=4502#a2}{https://www23.statcan.gc.ca/imdb/p2SV.pl?...}
) \\
A: &  General Social Survey: Caregiving and Care Receiving, Public Use Microdata File (\href{https://www150.statcan.gc.ca/n1/en/catalogue/89M0031X}{https://www150.statcan.gc.ca/...}) \\
U: &  I'm sorry, but im having trouble finding the download link for cycle 32 (2018) \\
A: &  you have to click on 'More information' and that will bring you to the order form \\
U: &  ohhhh, I see \\
U: &  thank you for letting me know \\
U: &  thank you for your help! \\
A: &  you're welcome - happy to help! \\
\midrule
& \textbf{Out of domain requests}\\
U: &  I'm wondering how to obtain a copy of an autopsy that's been done for a family member. Doctor office says they don't get the autopsy report. \\
A: &  Statistics Canada does not issue certificates for birth, death or marriage. Civil registration is a provincial jurisdiction. Therefore, you must contact the government ministry, registry or vital statistics office of the province or territory in which the birth, death or marriage occurred. To obtain the contact information for your provincial or territorial government, please call Service Canada at 1-800-O-Canada (1-800-622-6232) or visit the following page:  Vital Statistics contacts (\href{https://www.statcan.gc.ca/eng/health/vital/2012001/con}{https://www.statcan.gc.ca/eng/health...}) . \\
A: &  also: Coroner's report (\href{https://spvm.qc.ca/en/Fiches/Details/Coroners-report}{https://spvm.qc.ca/en/...})\\
\bottomrule
\end{tabular}
    \caption{Examples of user intents described in \autoref{sec:userintents}.}
    \label{tab:examples_user_intents}
\end{table*}

\begin{table*}[h]
    \small
    \centering
    \begin{tabular}{l p{0.94\linewidth}}
    \hline\\
    U:  & Hi, I'm looking to obtain quarterly data in regards to GDP grow (Canada), BC Housing STarts, Canada Oil Price/BBL\\
    A: & Hello, my name is Kelly C. Give me one moment as I search\\
    \\
    U: & thanks! \\
    A: & For GDP growth rates, please consult the following link: Gross domestic product (GDP) at basic prices, by industry, monthly, growth rates\\
    A: & I'll continue searching for the rest. One moment \\
    A: & What do you mean by BC Housing Starts? \\
    \\
    U: & I'm required to research all of the housing starts for BC on a quarterly basis\\
    A: & Define "starts".\\
    \\
    U: & unit basis\\
    U: & Housing starts are the number of new residential construction projects that have begun during any particular month\\
    A: & Perfect. Give me one moment.\\
    A: & I'm still searching. Sorry about the wait.\\
    \\
    U: & no worries\\
    A: & I would have monthly data regarding new building permits being issued.\\
    \\
    U: & I'll take a look at that\\
    A: & Building permits, by type of structure and type of work (\href{https://www150.statcan.gc.ca/t1/tbl1/en/cv.action?pid=3410006601}{https://www150.statcan.gc.ca/t1/...})\\
    A: & Do play with the variables to get what you are looking for.\\
    \\
    U: & thank you\\
    A: & To view/ manipulate the variables available in a data table: • Click on "Add/Remove data" • Select at least one variable in each tab and click on “Apply” in order to view your customized table To download the data: • Click on “Download” • Select one of the three output formats then click on the hyperlink and save the table The following video may also be helpful to you: How to use the data tables (\urlx{https://www.statcan.gc.ca/eng/sc/video/howto}) .\\
    A: & I'll have a look for oil prices. One moment.\\
    \\
    U: & Do you also have data to Canada'as oil Price/BBL ("WTI")?\\
    U: & okay thanks!\\
    A: & Are you looking for the retail prices of oil?\\
    A: & If so, I found some data for smaller geographies.\\
    A: & Monthly average retail prices for gasoline and fuel oil, by geography (\href{https://www150.statcan.gc.ca/t1/tbl1/en/cv.action?pid=1810000101}{https://www150.statcan.gc.ca/t1/...})\\
    \\
    U: & yes\\
    A: & Would those geographies be enough?\\
    A: & Or are you looking for Canada only?\\
    \\
    U: & hmm\\
    U: & I would need something that pertains more to all of canada\\
    A: & What about this? Monthly average retail prices for food and other selected products (\href{https://www150.statcan.gc.ca/t1/tbl1/en/cv.action?pid=1810000201}{https://www150.statcan.gc.ca/t1/...})\\
    A: & Is there anything else I can do for you?\\
    \\
    U: & I thinks that's all\\
    U: & I\\
    U: & i'll try and use this thanks!\\
    A: & No problem. Have a good one!\\
    \hline
    \end{tabular}
    \caption{Full Example (Sample \#42) taken from the conversations dataset. URLs were updated to link to default view (rather than filtered). See \autoref{tab:selected_example_conversation} for truncated conversation.}
    \label{tab:full_example_conversation}
\end{table*}

\begin{figure*}[h]
    \small
    \centering
    \includegraphics[width=0.90\linewidth]{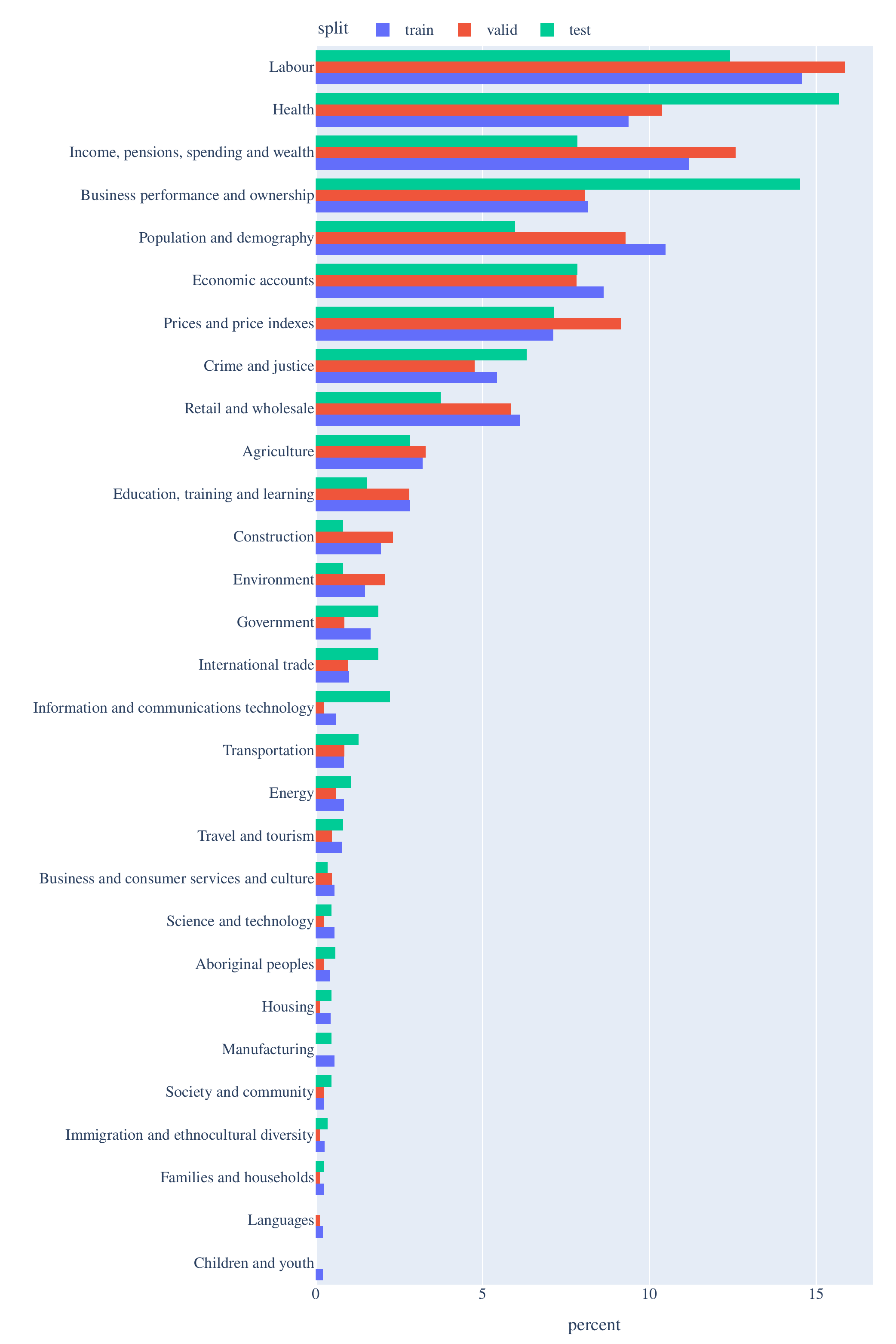}
    \caption{Distribution of subject categories for tables in the retrieval task, colored by the split.}
    \label{fig:category_distribution}
\end{figure*}

\begin{table*}[h]
    \small
    \centering
    \begin{tabular}{p{0.9\linewidth}}

\toprule

Title: Production and value of maple products\\

\midrule

Date range: 1924-01-01 to 2020-01-01\\
Dimensions: Geography, Maple products\\
Subject: Agriculture\\
Survey: Maple Products\\
Frequency: Annual\\

\midrule

Geography:\\
ID: 1, Parent: None, Name: Canada\\
ID: 2, Parent: 1, Name: Nova Scotia\\
ID: 3, Parent: 1, Name: New Brunswick\\
ID: 4, Parent: 1, Name: Quebec\\
ID: 5, Parent: 1, Name: Ontario\\
Maple products:\\
ID: 9, Parent: None, Name: Maple products expressed as syrup, total\\
ID: 10, Parent: None, Name: Gross value of maple products\\
ID: 1, Parent: None, Name: Production of maple sugar\\
ID: 2, Parent: None, Name: Value of maple sugar\\
ID: 3, Parent: None, Name: Production of maple syrup\\
ID: 4, Parent: None, Name: Value of maple syrup\\
ID: 5, Parent: None, Name: Production of maple taffy\\
ID: 6, Parent: None, Name: Value of maple taffy\\
ID: 7, Parent: None, Name: Production of maple butter\\
ID: 8, Parent: None, Name: Value of maple butter\\

\midrule

ID: 0, Note: Conversion factors: 1 gallon of syrup equals 10.0 pounds of maple sugar. One gallon of syrup weighs 13.24760 pounds. One gallon of syrup equals 10.4 pounds of taffy. Maple taffy is reported by Quebec and Nova Scotia only and commenced reporting in 1965 and 1983 respectively. The conversion of maple taffy to syrup varies with the density of syrup that year.\\
ID: 0, Note: Commercial production and value figures exclude inventory.\\
ID: 2, Note: Estimates produced by Nova Scotia horticulture industry specialists.\\
ID: 3, Note: Beginning in 1986, survey data for New Brunswick and Ontario are weighted using the number of taps made on maple trees in the spring of the year of the most recently completed census of agriculture.\\
ID: 4, Note: Estimates produced by l'Institut de la statistique du Québec, Groupe AGÉCO and Fédération des producteurs acéricoles du Québec.\\
ID: 5, Note: Beginning in 1986, survey data for New Brunswick and Ontario are weighted using the number of taps made on maple trees in the spring of the year of the most recently completed census of agriculture.\\
ID: 9, Note: “These products are represented by North American Product Classification System (NAPCS) Canada 2017 category 115136 \\

\hline
\hline

Maple syrup and other maple products. Data collected for maple butter, maple taffy and maple sugar are converted into a maple syrup equivalent and are included in this category.”\\
ID: 10, Note: “These products are represented by North American Product Classification System (NAPCS) Canada 2017 category 115136 – Maple syrup and other maple products. Data collected for maple butter, maple taffy and maple sugar are converted into a maple syrup equivalent and are included in this category.”\\
ID: 0, Note: “These products are represented by North American Product Classification System (NAPCS) Canada 2017 category 115136 – Maple syrup and other maple products. Data collected for maple butter, maple taffy and maple sugar are converted into a maple syrup equivalent and are included in this category.”\\
\bottomrule
    \end{tabular}
    \caption{Full metadata of table in \autoref{sec:table_specs}. Single lines were added to delimit the scope: Title, basic information, item names, and footnotes. The double line was added to delimit the truncation limit of the DPR and TAPAS model (512 tokens). The basic information is presented in \autoref{tab:sample_basic_info}.}
    \label{tab:sample_full_info}
\end{table*}

\begin{table}[h]
    \small
    \centering
    \begin{tabular}{l p{0.75\linewidth}}
\toprule
\textbf{PID} & \textbf{18100004 (count = 170)} \\
Title &  Consumer Price Index, monthly, not seasonally adjusted \\
Dates &  1914-01-01 to 2021-08-01 \\
Dim. &  Geography, Products and product groups \\
Subject &  Prices and price indexes \\
Survey &  Consumer Price Index \\
Freq. &  Monthly \\

\midrule
\textbf{PID} & \textbf{11100222 (count = 125)} \\
Title &  Household spending, Canada, regions and provinces \\
Dates &  2010-01-01 to 2019-01-01 \\
Dim. &  Geography, Statistic, Household expenditures, summary-level categories \\
Subject &  Income, pensions, spending and wealth \\
Survey &  Survey of Household Spending \\
Freq. &  Annual \\

\midrule
\textbf{PID} & \textbf{17100005 (count = 123)} \\
Title &  Population estimates on July 1st, by age and sex \\
Dates &  1971-01-01 to 2021-01-01 \\
Dim. &  Geography, Sex, Age group \\
Subject &  Population and demography \\
Survey &  Annual Demographic Estimates Canada, Provinces and Territories \\
Freq. &  Annual \\

\midrule
\textbf{PID} & \textbf{33100214 (count = 117)} \\
Title &  Canadian Business Counts, with employees, June 2019 \\
Dates &  2019-01-01 to 2019-01-01 \\
Dim. &  Geography, Employment size, North American Industry Classification System (NAICS) \\
Subject &  Business performance and ownership \\
Survey &  Business Register \\
Freq. &  Semi-annual \\

\midrule
\textbf{PID} & \textbf{35100177 (count = 104)} \\
Title &  Incident-based crime statistics, by detailed violations, Canada, provinces, territories and Census Metropolitan Areas \\
Dates &  1998-01-01 to 2020-01-01 \\
Dim. &  Geography, Violations, Statistics \\
Subject &  Crime and justice \\
Survey &  Uniform Crime Reporting Survey \\
Freq. &  Annual \\

\midrule
\textbf{PID} & \textbf{18100005 (count = 91)} \\
Title &  Consumer Price Index, annual average, not seasonally adjusted \\
Dates &  1914-01-01 to 2020-01-01 \\
Dim. &  Geography, Products and product groups \\
Subject &  Prices and price indexes \\
Survey &  Consumer Price Index \\
Freq. &  Annual \\

\bottomrule

\end{tabular}
    \caption{Most frequently retrieved tables across all splits in the retrieval task. The \textit{basic information} is provided for each table. Analysis presented in \autoref{sec:appendix_frequently_requested_tables}.}
    \label{tab:most_popular_tables}
\end{table}

\begin{table}[h]
    \small
    \centering
    \begin{tabular}{l p{0.85\linewidth}}
\toprule

U: & hi, i was wondering if you vae any statistics on video game sales, ot high school drop out rates? {\color{blue} infoSeekingQuestions; request;} 
\\ A: & Hello my name is Sylvain, how may I help you? {\color{blue} offer;}
\\ U: & hello, I am looking for high school drop out r ates and teen depression rates, but cant seem to find anything that goes back more than a year {\color{blue} request;}
\\ A: & Data for High School dropouts is compiled by the Provincial Education Ministry. To obtain this infor please contact the Education Ministry of your province {\color{blue} inform; promise;}
\\ A: & Data for mood disorder (depression) is available on our website from the Data Table 13- 10-0096-18 (\href{https://www150.statcan.gc.ca/t1/tbl1/en/tv.action?pid=1310009618}{https://www150.statcan.gc.ca/...}) {\color{blue} answer;}
\\ A: & To view/ manipulate the variables available in a data table: • Click on "Add/Remove data" • Select at least one variable in each tab and click on “Apply” in order to view your customized table To download the data: • Click on “Download” • Select one of the three output formats then click on the hyperlink and save the table The following video may also be helpful to you: How to use the data tables (\href{https://www.statcan.gc.ca/eng/sc/video/howto}{https://www.statcan.gc.ca/eng/sc/video/...}). {\color{blue} inform; request;}
\\

\midrule

U: &  Hi team, can you please help me retrieve data of public deposits of chartered banks? \textcolor{blue}{infoSeekingQuestions; request; }
\\ A: &  Please hold \textcolor{blue}{timeManagement; }
\\ U: &  Surely. \textcolor{blue}{autoFeedback; }
\\ A: &  You can obtain this info on our website from our Data Table 10-10-0116-01 (\href{https://www150.statcan.gc.ca/t1/tbl1/en/tv.action?pid=1010011601}{https://www150.statcan.gc.ca/t1/...}) \textcolor{blue}{answer;}
\\ A: &  To view/ manipulate the variables available in a data table: • Click on "Add/Remove data" • Select at least one variable in each tab and click on ``Apply" in order to view your customized table To download the data: • Click on ``Download" • Select one of the three output formats then click on the hyperlink and save the table The following video may also be helpful to you:  How to use the data tables (\href{https://www.statcan.gc.ca/eng/sc/video/howto}{https://www.statcan.gc.ca/eng/sc/video/...}) . \textcolor{blue}{inform; instruct; }
\\ A: &  Please take note that this is a non-Statistics Canada Table. This info is provide from The Bank of Canada \textcolor{blue}{inform; }
\\ U: &  Thanks a lot! So, how should I get down to Public deposits? or rather which of these lines would direct me to Public deposits? \textcolor{blue}{clarificationQuestion; }
\\ A: &  Under the category Chartered bank deposits. For more info please contact The Bank of Canada on their website by using their Online Request Form (\href{https://www.bankofcanada.ca/about/contact-information/general-inquiries/#online}{https://www.bankofcanada.ca/...}) \textcolor{blue}{answer; inform; }
\\
\bottomrule
\end{tabular}
    \caption{Samples of the dialogue analysis in \autoref{sec:dialogue_analysis}, which are summarized in \autoref{tab:iso_speech_acts_count}. The speech acts are in \textcolor{blue}{blue} at the end of each message. The remaining annotations are in the supplementary materials. Sample examples are given in \autoref{tab:merged_speech_act_counts}.}
    \label{tab:sample_linguistic_analysis}
\end{table}

\begin{table*}[h]
    \small
    \centering
    \begin{tabular}{l p{0.85\linewidth}}
    \toprule
    \textbf{U:} & \textbf{Thank you for your information, but I have done that already. Your previous website had a "manipulate" function to calculate the percentage change. I don't see the button anymore, and wondered if you have other schedule prepared as that function is gone.}
    \\
    \midrule
    \textbf{A:} & \textbf{Hello, my name is Jimmy N., how may I help you?}
    \\
    T5: & Hello, my name is Jimmy N., how may I help you?
    \\
    T5+D: & Hello, my name is Jimmy N., how may I help you?
    \\
    \midrule
    \textbf{U:} & \textbf{I was talking with Sylvain, will you be able to see the log?}
    \\
    \midrule
    \textbf{A:} & \textbf{Yes, please hold while I verify.}
    \\
    T5: & Yes, please hold while I verify.
    \\
    T5+D: & Yes, please hold while I verify.
    \\
    \midrule
    \textbf{A:} & \textbf{Just to clarify, you're looking for all-items ?}
    \\
    T5: & You can obtain this information at the following hyperlink : Table: 18-10-0004-13 Consumer Price Index by product group, monthly, percentage change, not seasonally adjusted, Canada, provinces, Whitehorse, Yellowknife and Iqaluit ( )
    \\
    T5+D: & You can obtain this information at the following hyperlink : Table: 18-10-0004-01 Consumer Price Index by product group, monthly, [...] and Iqaluit ( )
    \\
    \bottomrule
    \end{tabular}
    \caption{Conversation \#4890. Each message is separated by a horizontal line. The original conversation, in \textbf{bold}, only contains messages by the user (U) and agent (A). The generated responses by T5 and T5 with top-5 DPR-retrieved titles (T5+D) predict the corresponding agent's message (in \textbf{bold}).}
    \label{tab:sample_conversation_4890}
\end{table*}

\begin{table*}[h]
    \small
    \centering
    \begin{tabular}{l p{0.85\linewidth}}
        \toprule
         & [...]
        \\
        \textbf{U:} & \textbf{Ideally, I want the percentage change for these items}
        \\
        $R_1$: & \textit{Consumer Price Index, monthly, not seasonally adjusted }
        \\
        $R_2$: & \textit{Monthly adjusted price index, provisional}
        \\
        $R_3$: & \textit{Consumer Price Index, monthly, seasonally adjusted}
        \\
        $R_4$: & \textit{Historical (real-time) releases of the industrial product price index, by major product group, monthly, inactive}
        \\
        $R_5$: & \textit{Consumer Price Index (CPI) statistics, alternative measures, unadjusted and seasonally adjusted, Bank of Canada}
        \\
        \midrule
        \textbf{A}: & \textbf{You can obtain this information at the following hyperlink : Table: 18-10-0004-13 Consumer Price Index by product group, monthly, percentage change, not seasonally adjusted, Canada, provinces, Whitehorse, Yellowknife and Iqaluit (\href{https://www150.statcan.gc.ca/t1/tbl1/en/tv.action?pid=1810000413}{https://www150.statcan.gc.ca/t1/tbl1/...})}
        \\
        T5: & You can obtain this information at the following hyperlink : Table: 18-10-0004-13 Consumer Price Index [...] and Iqaluit ( )
        \\
        T5+D: & You can obtain this information at the following hyperlink : Table: 18-10-0004-13 Consumer Price Index [...] and Iqaluit ( )
        \\
        \bottomrule
    \end{tabular}
    \caption{The second part of conversation \#4890, immediately after \autoref{tab:sample_conversation_4890}. $R_i$ indicates the $i$-th table retrieved by DPR, which is only seen by T5+D.}
    \label{tab:sample_conversation_4890_part_2}
\end{table*}

\begin{table*}[]
    \small
    \centering
    \begin{tabular}{l p{0.85\linewidth}}
\toprule
U: & This is pretty cool! 
\\
A: & Hello, my name is Martine, how may I help you? 
\\
U: & I have a question - how would I be able to access data for 2019 on the amount of money the federal government spent (money out) to Canadians through social programs like EI / OAS etc... 
\\
A: & Letme check.. 
\\
U: & Thank you
\\
$R_i$: & \textit{Federal extramural expenditures on science and technology, type of science, performing sector, type of payment and geography}
\\
$R_i$: & \textit{Historical: Current transfers to other levels of government, 1968 System of National Accounts (SNA), annual, 1926 - 1986}
\\
$R_i$: & \textit{Private non-profit organization outsourced research and development expenditures}
\\
$R_i$: & \textit{Capital expenditures, sources of funding, by geography}
\\
$R_i$: & \textit{Historical: Government transfer payments to persons, 1968 System of National Accounts (SNA), annual, 1926 - 1986}
\\
\midrule
\textbf{A}: & \textbf{Revenue, expenditure and budgetary balance - General governments (x 1,000,000) (\href{https://www150.statcan.gc.ca/t1/tbl1/en/tv.action?pid=3610047701}{https://www150.statcan.gc.ca/t1/...3610047701})}
\\
T5: & Data Tables (\urlx{https://www150.statcan.gc.ca/n1/en/type/data?text=federal+debt})
\\
T5+D: & Federal extramural expenditures on science and technology, type of science, performing sector, type of payment and geography ( )
\\
\bottomrule
    \end{tabular}
    \caption{Conversation \#8960.}
    \label{tab:sample_conversation_8960}
\end{table*}

\begin{table*}[]
    \small
    \centering
    \begin{tabular}{l p{0.85\linewidth}}
\toprule

U: & How can I find information relating to Kelowna population estimates, demography, population projection estimates, population by age and sex 
\\
A: & hello how may i help you today? 
\\
U: & How can I find information relating to Kelowna population estimates, demography, population projection estimates, population by age and sex 
\\
A: & let me see what i can find on the subject for the projections information our tables only provide it to the provincial level it does not go lowe then that  <extra\_id\_0>  ( <extra\_id\_0> )
\\
$R_1$: & \textit{Estimates and projections of population aged 0 to 29, by age group, Canada, provinces and territories}
\\
$R_2$: & \textit{Projected population, by projection scenario, age and sex, as of July 1}
\\
$R_3$: & \textit{Population estimates on July 1st, by age and sex}
\\
$R_4$: & \textit{Components of projected population growth, by projection scenario}
\\
$R_5$: & \textit{Population estimates, quarterly}
\\
\midrule
\textbf{A}: & \textbf{\urlx{https://www150.statcan.gc.ca/t1/tbl1/en/cv.action?pid=1710005801}}
\\
T5: & \urlx{https://www150.statcan.gc.ca/n1/daily-quotidien/190619/dq190619f-cansim-eng.htm}
\\
T5+D: & \urlx{https://www150.statcan.gc.ca/n1/pub/71-607-x/71-607-x2018005-eng.htm}
\\
\bottomrule
    \end{tabular}
    \caption{Conversation \#21533.}
    \label{tab:sample_conversation_21533}
\end{table*}

\begin{table*}[h]
    \small
    \centering
    \begin{tabular}{l p{0.9\linewidth}}
        \toprule
        U: & Hi there, do you know approximately how many craft distillers are operating in Canada?
        \\
        A: & Hello my name is Sylvain, how may I help you? 
        \\
        U: & I was wondering if you had any information about craft distillers in Canada. Specifically, how many craft distillers are operating in Can
        \\
        A: & The industries are classified according to the North American Industry Classification System (NAICS). To find the NAICS code for an indu refer to this page (\href{http://www23.statcan.gc.ca/imdb/p3VD.pl?Function=getVD&TVD=1181553}{http://www23.statcan.gc.ca/imdb/...1181553}) . You may start your search by selecting one of the 20 sectors, until a six-d obtained. [...] You will find the number of business locations by NAICS code in the following data tables: Table 33-10-0105-01 Canadian Business Counts, with employees, December 2018 (\href{https://www150.statcan.gc.ca/t1/tbl1/en/tv.action?pid=3310010501}{https://www150.statcan.gc.ca/t1/...3310010501}) [...] I also suggest consulting the financial data of small and medium enterprise in Canada; these organizations have annual earning revenues ranges of 
        \$30,000 to \$20 million. T information is available by industry, province, territory and legal status of companies (incorporated or unincorporated). This data is not available on our website; however, i available on Innovation, Science and Economic Development Canada’s website, using their Financial Performance Data (\href{http://www.ic.gc.ca/eic/site/pp-pp.nsf/eng/home}{http://www.ic.gc.ca/...})[...]
        \\
        \bottomrule

    \end{tabular}
    \caption{Conversation \#1628}
    \label{tab:sample_conversation_1628}
\end{table*}

\begin{table*}[h]
    \small
    \centering
    \begin{tabular}{l p{0.85\linewidth}}
    \toprule
    U: & Hi. I want to know how many small, medium and big companies are there in canada. Thank you
    \\
    A: & Hello, my name is Olivier C. Please wait while i find the information you're looking for.
    \\
    
    $R_1$: & \textit{Canadian Business Counts, without employees, December 2019}
    \\
    $R_2$: & \textit{Canadian Business Counts, with employees, December 2019}
    \\
    $R_3$: & \textit{Canadian Business Counts, without employees, June 2019}
    \\
    $R_4$: & \textit{Canadian Business Counts, with employees, June 2019}
    \\
    $R_5$: & \textit{Canadian Business Counts, without employees, December 2017}
    \\
    \midrule
    \textbf{A}: & \textbf{Thank you for waiting, you will find the information about the number of business per business size if you click the following hyperlinks : Canadian Business Counts, with employees, December 2019 (\href{https://www150.statcan.gc.ca/t1/tbl1/en/tv.action?pid=3310022201}{https://www150.statcan.gc.ca/t1/...pid=3310022201}) Canadian Business Counts, without employees, December 2019 (\href{https://www150.statcan.gc.ca/t1/tbl1/en/tv.action?pid=3310022301}{https://www150.statcan.gc.ca/t1/...pid=3310022301})}
    \\
    T5: & Thank you for waiting. You will find the information you're looking for if you click the following hyperlink : Canadian Business Counts, with employees, June 2019 ( )
    \\
    T5+D: & Thank you for waiting. The industries are classified according to the North American Industry Classification System (NAICS). To find the NAICS code for an industry, please refer to this page (\href{http://www23.statcan.gc.ca/imdb/p3VD.pl?Function=getVD&TVD=1181553}{http://www23.statcan.gc.ca/imdb/...1181553}). You may start your search by selecting one of the 20 sectors, until a six-digit NAICS is obtained. You will find the number of business locations by NAICS code in the following data tables: Table 33-10-0214-01 Canadian Business Counts, with employees, June 2019 ( ) Table 33-10-0215-01 Canadian Business Counts, without employees, June 2019 ( ) I also suggest consulting the financial data of small and medium enterprise in Canada; these organizations have annual earning revenues ranges of \$30,000 to \$20 million. This information is available by industry, province, territory and legal status of companies (incorporated or unincorporated). This data is not available on our website; however, it is freely available on Innovation, Science and Economic Development Canada's website, using their Financial Performance Data (\urlx{http://www.ic}
    \\
    \bottomrule
    \end{tabular}
    \caption{Conversation \#8500.}
    \label{tab:sample_conversation_8500}
\end{table*}

\begin{table*}[h]
    \small
    \centering
    \begin{tabular}{l p{0.85\linewidth}}
        \toprule
        U: & Do you have any data showing labour in manufacturing companies and how it is allocated between direct and indirect labour? 
        \\
        A: & Hello, my name is Jimmy N. Could you please specify what you mean by direct and indirect labour. 
        \\
        U: & Direct = labour for building product Indirect = administrative, supervision, etc. 
        \\
        A: & The data you are requesting is not available as a standard product. However, it may be available through a custom order (a fee may apply). [...] Would you like to make this request? 
        \\
        A: & As a standard product, we have tables about employment by industry.
        \\
        U: & I would just like to look at the standard product. Can you tell me where I can find it, or what it is called?\
        \\
        $R_1$: & \textit{Labour force characteristics by industry, monthly, seasonally adjusted, last 5 months}
        \\
        $R_2$: & \textit{Employment by industry, monthly, seasonally adjusted and unadjusted, and trend-cycle, last 5 months}
        \\
        $R_3$: & \textit{Employment by industry, annual}
        \\
        $R_4$: & \textit{Employment by industry, three-month moving average, unadjusted for seasonality, provinces and economic regions, inactive}
        \\
        $R_5$: & \textit{Usual hours worked by industry, annual}
        \\
        \midrule
        \textbf{A}: & \textbf{You can obtain this information at the following hyperlinks : Table: 14-10-0355-02 Employment by industry, monthly, seasonally adjusted (x 1,000)Table: 14-10-0023-01 Labour force characteristics by industry, annual (x 1,000)Employed labour force who worked full year, full time and reported employment income in 2015, 2016 Census (\href{https://www150.statcan.gc.ca/t1/tbl1/en/tv.action?pid=1410035502}{https://www150.statcan.gc.ca/t1/...1410035502})}
        \\
        T5: & Table: 14-10-0296-01 Labour force characteristics by industry, annual (x 1,000) ( )
        \\
        T5+D: & Please hold while I find the information.
        \\
        \bottomrule
    \end{tabular}
    \caption{Conversation \#19568.}
    \label{tab:sample_conversation_19568}
\end{table*}

\begin{table*}[]
\small
\centering
\begin{tabular}{l p{0.85\linewidth}}
\toprule
U: & I am on webpage \urlx{https://www150.statcan.gc.ca/n1/daily-quotidien/201224/dq201224beng.htm} where can I find the same data set for previous years, for example 2019 2018 2017 etc 
\\
A: & Hello, can you tell me on which table you are : \urlx{https://www150.statcan.gc.ca/n1/daily-quotidien/201224/dq201224b-cansim-eng.htm}
\\
U: & Provisional death counts and excess mortality, January to October 2020 Adjusted number of deaths, expected number of deaths and estimates of excess mortality, by week 
\\
A: & That is the report, but there is 5 different tables: \urlx{https://www150.statcan.gc.ca/n1/daily-quotidien/201224/dq201224b-cansim-eng.htm} Let me verify this for you. One moment please. To view/ manipulate the variables available in a data table: • Click on "Add/Remove data" • [...] Select at least one variable in each tab and click on “Apply” in order to view your customized table Data are available for 2016 up to 2020. 
\\
U: & The reference period stops at January 2020 I can't seem to get data for past years 
\\
A: & Click on the schedule icon and select the year you want. (2016 - 2020) 
\\
U: & I did and all years are 'greyed out' meaning they do not populate 
\\
A: & Let me verify this. One moment please. 
\\
U: & screenshot.lzsc
\\
$R_1$: & \textit{Weekly death counts, by age group and sex}
\\
$R_2$: & \textit{Previous releases and revisions to weekly death counts}
\\
$R_3$: & \textit{Adjusted number of deaths, expected number of deaths and estimates of excess mortality, by week, age group and sex}
\\
$R_4$: & \textit{Death counts, by week, inactive}
\\
$R_5$: & \textit{Labour force characteristics by sex and detailed age group, monthly, unadjusted for seasonality}
\\
\midrule
\textbf{A}: & \textbf{Have you already consulter the table that replaces table\# 13-10-0785-01 [...] \urlx{https://www150.statcan.gc.ca/t1/tbl1/en/tv.action?pid=1310081001} - On this table you can go back to 2016.}
\\
T5: & You can manipulate the table by following these steps: • Click on "Add/Remove data" • Click on the “ + ” icon next to a checkbox to expand the list of variables (if applicable) • Select at least one variable in each tab and click on “Apply” in order to view your customized table To download the data: • Click on “Download” • Select one of the three output formats then click on the hyperlink and save the table The following video may also be helpful to you: How to use the data tables (\urlx{https://www.statcan.gc.ca/eng/sc/video/howto}).
\\
T5+D: & Please consult the following data table(s): Weekly death counts, by age group and sex ( )
\\
\bottomrule
\end{tabular}
\caption{Conversation \#15376.}
\label{tab:sample_conversation_15376}
\end{table*}

\begin{table*}[]
\small
\centering
\begin{tabular}{l p{0.85\linewidth}}
\toprule
U: & Hello 
\\
A: & Hello, my name is Jimmy N., how may I help you? 
\\
U: & [...] We saw this article from the Vancouver Sun which mentioned StatsCan data on business closures. [...] is there any way for us to collect localized data for the Township, or at least the Fraser Valley 
\\
A: & Unfortunately, the information is available on a national and provincial/territorial level : And for some Census metropolitan areas
\\
$R_i$: & \textit{Experimental estimates for business openings and closures for Canada, provinces and territories, census metropolitan areas, seasonally adjusted}
\\
$R_i$: & \textit{Quarterly estimates of business entry and exit}
\\
$R_i$: & \textit{Of businesses where at least one employee was laid off, percentage of workforce laid off and rehired due to COVID-19, by business characteristics}
\\
$R_i$: & \textit{Canadian Business Counts, with employees, census metropolitan areas and census subdivisions, December 2020}
\\
$R_i$: & \textit{Manufacturing sales, by industry for 12 Selected Census Metropolitan Areas}
\\
\midrule
\textbf{A}: & \textbf{Table: 33-10-0270-01 - Experimental estimates for business openings and closures for Canada, provinces and territories, census metropolitan areas, seasonally adjusted (\href{https://www150.statcan.gc.ca/t1/tbl1/en/tv.action?pid=3310027001}{https://www150.statcan.gc.ca/t1/...3310027001})}
\\
T5: & However, it may be available through a custom order (a fee may apply). To obtain a free cost estimate for this service, please send us your request, clearly indicating the variables, geographic areas, time frame and data frequency needed, making sure to include your complete contact information (name of organization, address, postal code, and phone number) at the following email address: infostats@canada.ca (mailto:infostats@canada.ca) or by telephone at 1-800-263-1136. Our agents are available Monday to Friday (except holidays) from 7:30 am to 7:30 pm (Eastern Time). An agent will get in touch with you to verify the information, provide an estimate and proceed with the payment, if applicable.
\\
T5+D: & However, it may be available through a custom order (a fee may apply). To obtain a free [...] An agent will get in touch with you to verify the information, provide an estimate and proceed with the payment, if applicable. Please do not share any personal information through the live chat. Would you like to make this request?
\\
\bottomrule
\end{tabular}
\caption{Conversation \#11869.}
\label{tab:sample_conversation_11869}
\end{table*}

\begin{table*}[]
    \small
    \centering
\begin{tabular}{l p{0.6\linewidth} rrrr}
\toprule
PID &                                                                                                                                                 Title &  Train &  Valid &  Test &  Total \\
\midrule
13100810 &                                                                                                             Selected grouped causes of death, by week &      0 &      0 &     6 &      6 \\
33100270 &       Experimental estimates for business openings and closures for Canada, provinces and territories, census metropolitan areas, seasonally adjusted &      0 &      1 &     3 &      4 \\
13100763 &                                    Health characteristics of children and youth aged 1 to 17 years, Canadian Health Survey on Children and Youth 2019 &      0 &      0 &     4 &      4 \\
35100060 &                                              Number of homicide victims and persons accused of homicide, by Indigenous identity, age group and gender &      0 &      1 &     2 &      3 \\
32100042 &  Estimated areas, yield, production of corn for grain and soybeans, using genetically modified seed, Quebec and Ontario, in metric and imperial units &      0 &      1 &     2 &      3 \\
14100084 &                               Labour force characteristics of immigrants by sex and age group, three-month moving average, unadjusted for seasonality &      0 &      0 &     3 &      3 \\
13100143 &                                                                            Deaths, by cause, Chapter V: Mental and behavioural disorders (F00 to F99) &      0 &      0 &     3 &      3 \\
\bottomrule
\end{tabular}

    \caption{List of tables that do not appear in the training set, but appear at least 3 times in the validation or test sets.}
    \label{tab:tables_not_in_train_with_three_appearances}
\end{table*}

\begin{table*}[h]
    \small
    \centering
\begin{tabular}{lllrrrrr}
\toprule
Language & Split & Augmentation &  METEOR &  ROUGE-L &  BERTScore &  MoverScore &  Title Acc. \\
\midrule
English & Test & None &   23.35 &    30.65 &      86.04 &       59.82 &        6.96 \\
       &       & Top-1 Title &   24.07 &    30.76 &      86.11 &       60.23 &        7.99 \\
       &       & Top-5 Titles &   24.41 &    30.88 &      86.17 &       60.31 &       10.82 \\
       & Valid & None &   31.95 &    43.31 &      88.31 &       65.50 &       30.39 \\
       &       & Top-1 Title &   33.11 &    44.22 &      88.60 &       66.14 &       33.66 \\
       &       & Top-5 Titles &   33.38 &    44.20 &      88.55 &       65.94 &       38.89 \\
\midrule
French & Test & None &    5.98 &     7.21 &      61.74 &       51.88 &        0.00 \\
       &       & Top-1 Title &    9.71 &    11.14 &      65.86 &       54.81 &        0.00 \\
       &       & Top-5 Titles &    8.96 &    10.75 &      64.76 &       54.38 &        0.00 \\
       & Valid & None &    6.84 &     8.85 &      60.97 &       52.21 &        0.00 \\
       &       & Top-1 Title &    8.09 &     9.03 &      64.22 &       53.62 &        0.00 \\
       &       & Top-5 Titles &    7.37 &     8.75 &      62.55 &       52.97 &        0.00 \\
\bottomrule
\end{tabular}
    \caption{Full response generation results. Selected results in \autoref{tab:selected_generation_results_english}.}
    \label{tab:full_generation_results}
\end{table*}

\end{document}